\documentclass[a4,dvipsnames]{article}
\usepackage{graphicx}
\usepackage{amsmath, latexsym, amssymb, bm, ascmac, here, mathtools, amsthm, subfig}
\usepackage{xcolor}
\usepackage{here}
\usepackage{authblk}
\usepackage{natbib}
\usepackage{multirow}
\usepackage{arydshln}
\usepackage[ruled,linesnumbered,vlined]{algorithm2e}
\usepackage{url}
\usepackage{color, hyperref}
\usepackage{cases}
\usepackage{empheq}

\def\*#1{\boldsymbol{#1}}

\newtheorem{theorem}{Theorem}[section]

\newtheorem{lemma}{Lemma}[section]

\newcommand{\eq}[1]{(\ref{#1})}

\newcommand{\lw}[1]{\smash{\lower2.ex\hbox{#1}}}

\newcommand{\RR}{\mathbb{R}}

\newcommand{\cB}{{\mathcal B}}
\newcommand{\cC}{{\mathcal C}}
\newcommand{\cD}{{\mathcal D}}

\newcommand{\cF}{{\mathcal F}}
\newcommand{\cG}{{\mathcal G}}
\newcommand{\cH}{{\mathcal H}}
\newcommand{\cI}{{\mathcal I}}

\newcommand{\cL}{{\mathcal L}}

\newcommand{\cS}{{\mathcal S}}

\newcommand{\cW}{{\mathcal W}}

\usepackage[margin=1in]{geometry}

\newcommand{\ed}[1]{#1}
\newcommand{\red}[1]{#1}
\newcommand{\blue}[1]{#1}
\newcommand{\yellow}[1]{\textcolor{yellow!60!red}{#1}}
\newcommand{\green}[1]{\textcolor{green!60!black}{#1}}
\newcommand{\B}[1]{$\pmb{#1}$}
 
\SetKwBlock{Function}{function}{end function}
\SetKwComment{CommentTmp}{\blue{$\triangleright$\:}}{}
\DontPrintSemicolon
\newcommand{\Comment}[1]{\CommentTmp*[r]{\small\blue{#1}\!\!\!\!\!\!}}
\newcommand{\CommentHere}[1]{\hfill\CommentTmp*[h]{\small\blue{#1}\!\!\!\!\!\!}}
\newcommand\addtag{\refstepcounter{equation}\tag{\theequation}}

\begin{document}

\title{Distance Metric Learning for \\
Graph Structured Data
}





\author[1]{Tomoki~Yoshida}
\author[1,2,3]{Ichiro~Takeuchi}
\author[1,2,4]{Masayuki~Karasuyama}

\affil[1]{Nagoya Institute of Technology}
\affil[2]{National Institute for Material Science}
\affil[3]{RIKEN Center for Advanced Intelligence Project}
\affil[4]{Japan Science and Technology Agency}
\affil[ ]{\textit{yoshida.t.mllab.nit@gmail.com, \{takeuchi.ichiro,karasuyama\}@nitech.ac.jp}}

\date{}

\maketitle

\begin{abstract}
 Graphs are versatile tools for representing structured data.
 \ed{As a result,}
 a variety of machine learning methods have been studied for graph data analysis.
 Although many \ed{such}
 learning methods depend on the measurement of differences between input graphs, defining an appropriate distance metric for \red{graphs} remains a controversial issue.
 %
 Hence, we propose a supervised distance metric learning method for the graph classification problem. 
 %
 Our method, named \emph{interpretable graph metric learning} (IGML), learns discriminative metrics in a subgraph-based feature space, which has a strong graph representation capability.
	%
	By introducing a sparsity-inducing penalty on \ed{the} weight of each subgraph, IGML can identify a small number of important subgraphs that can provide insight \ed{into} the given classification task.
	\blue{Because} our formulation has a large number of optimization variables, an efficient algorithm \ed{that uses} pruning techniques based on \emph{safe screening} and \emph{working set selection} methods \ed{is also proposed}.
	An important property of IGML is that solution \ed{optimality} is guaranteed because the problem is formulated as a convex problem and our pruning strategies only discard unnecessary subgraphs.
 %
 \ed{Furthermore}, we show that IGML is also applicable to other structured data such as \blue{itemset} and sequence data, and that it can incorporate vertex-label similarity by using a transportation-based subgraph feature.
	We empirically evaluate the computational efficiency and classification performance \ed{of IGML} on several benchmark datasets and \ed{provide} some illustrative examples \ed{of how} IGML identifies important subgraphs from a given graph dataset.
\end{abstract}

\section{Introduction}
\label{sec:intro}

Because of the growing diversity of data science applications, machine learning methods \ed{must} adapt to a variety of complicated structured data, from which it is often difficult to obtain \ed{typical} numerical vector representations of input objects.
A standard approach to modeling structured data is to employ \red{\emph{graphs}}.
%
For example, \ed{graph-based representations are prevalent} in domains such as chemo- and bio- informatics.
%
In this study, we particularly focus on the case in which a data instance is represented as a pair of a graph and \ed{its} associated class label.

%
Although numerous machine learning methods explicitly \ed{or} implicitly depend on how to measure differences between input objects, defining an appropriate distance metric on \red{graphs} remains a controversial issue in the \ed{field}.
%
A widely accepted approach is \ed{the} \emph{graph kernel} \citep{gartner2003graph,vishwanathan2010graph}, which enables to apply machine learning methods to graph data without \ed{requiring} explicit vector representations.
%
%
Another popular approach would be to use neural \ed{networks} \citep{atwood2016diffusion,narayanan2017graph2vec}, \ed{from} which a suitable representation \ed{can be} learned \ed{while avoiding to explicitly define a} metric.  
%
%
%
\red{
However, in these approaches, it is difficult to create a metric that explicitly extracts significant sub-structures, i.e., subgraphs.
Identifying discriminative subgraphs in an interpretable manner can be insightful for many graph classification tasks.
In particular, graph representation is prevalent in scientific data analysis.
For example, chemical compounds are often represented by graphs; thus, finding subgraphs that have a strong effect on a target label (e.g., toxicity) is informative.
Other examples of graph representations are protein 3D structures and crystalline substances \citep[e.g.,][]{brinda2005network,xie2018crystal}, where the automatic identification of important sub-structures is expected to provide an insight behind correlation between structures and target labels.
}
%
%
Further details of the \ed{previous} studies are discussed in Section~\ref{sec:related-work}.


We propose a supervised method \ed{that} obtains a metric for \red{graphs}, \ed{thereby} achieving both high predictive performance and interpretability.
%
Our method, named \emph{interpretable graph metric learning} (IGML), combines the concept of \emph{metric learning} \citep[e.g.,][]{weinberger2009distance,davis2007information} with a subgraph representation, where each graph is represented \ed{by} a set of \ed{its} subgraphs.
IGML optimizes a metric \ed{that assigns} a weight $m_{i(H)} \ge 0$ \ed{to} each subgraph $H$ contained in a given graph $G$.
%
Let $\phi_H(G)$ be a feature of the graph $G$ \ed{that} is monotonically non-decreasing with respect to the frequency of subgraph $H$ \ed{of} $G$.
Note that we assume that subgraphs are counted without overlapped vertices and edges throughout the study.
We consider the following squared distance between two graphs $G$ and $G'$:
\[ 
 d_{\*m}(G,G^\prime) 
 \coloneqq
 \sum_{H \in \cG}
 m_{i(H)}
 \left(
 \phi_H(G) - \phi_H(G^\prime) 
 \right)^2, 
 \addtag \label{eq:sq-distance}
\]
where $\cG$ is the set of all connected graphs.
%
Although it is known that the subgraph approach has strong graph representation capability \citep[e.g.][]{gartner2003graph}, na\"ive calculation is obviously infeasible \ed{unless} the weight parameters have some special structure.


We formulate IGML as a supervised learning problem of the distance function \eqref{eq:sq-distance} using a pairwise loss function of metric learning \citep{davis2007information} with a sparse penalty on $m_{i(H)}$.
The resulting optimization problem is computationally infeasible at a glance, because the number of weight parameters is equal to the number of possible subgraphs, which is usually intractable.
We overcome this difficulty by introducing \emph{safe screening} \citep{ghaoui2010safe} and \emph{working set selection} \citep{fan2008liblinear} \red{approaches}.
%
%
Both of these approaches can \ed{significantly} reduce the number of variables, and further, they can be combined with a \emph{pruning} strategy on the tree traverse of \emph{graph mining}.
These optimization tricks are inspired by two recent studies \citep{nakagawa2016safe} and \citep{morvan2018whinter}, which \ed{developed} safe screening- and working set- based pruning for a linear prediction model with the LASSO penalty, \ed{respectively}. 
%
By combining these two techniques, we \ed{constructed} a path-wise optimization method that can obtain \ed{a} sparse solution of the weight parameter $m_{i(H)}$ without directly enumerating all possible subgraphs. 

To \ed{the best of} our knowledge, no \ed{previous} studies can provide an interpretable subgraph-based metric \ed{learned} in a supervised manner.
%
%
The advantages of IGML can be summarized as follows:
\begin{itemize}
 \item \ed{Because} IGML is formulated as a convex optimization \red{problem}, the global optimal can be found by the standard gradient-based optimization.

 \item The safe screening- and working set-based optimization algorithms make our problem practically tractable without sacrificing optimality.
       %

 \item We can identify a small number of important subgraphs that discriminate different classes. 
       This \ed{implies} that the resulting metric is easy to compute and highly interpretable, \ed{making it} useful for a variety of subsequent data \red{analyses}.
       For example, applying the nearest neighbor classification or decision tree on the learned space would be effective.
\end{itemize}
%
\ed{Moreover,} we propose three extensions of IGML.
First, we show that IGML is directly applicable to other structured data, such as \ed{itemset} and sequence data.
Second, \ed{its} application to a triplet based loss function is discussed.
Third, we extend IGML \ed{to allow} similarity information of vertex-labels \ed{to} be incorporated.
%
\ed{We} empirically verify the superior or comparable prediction performance of IGML to other existing graph classification methods (most of which \ed{are not interpretable}).
We also show some examples of extracted subgraphs and data analyses on the learned metric space.

\ed{The reminder of} this paper is organized as follows. 
In Section~\ref{sec:related-work}, we review \ed{previous} studies on graph data analysis. 
In Section~\ref{sec:learning}, we introduce a formulation of our proposed IGML.
Section~\ref{sec:safe-and-working} discusses strategies \ed{to} reduce the size of the \ed{IGML} optimization problem.
The detailed computational procedure of IGML is described in Section~\ref{sec:computations}.
\ed{Three} extensions of IGML are presented in Section~\ref{sec:extension}.
Section~\ref{sec:experiments} \ed{reports our empirical evaluation of} the effectiveness of IGML on several benchmark datasets.

Note that this paper is an extended version of \ed{a} preliminary conference paper \citep{yoshida2019learning}.
The source code of the program used in \ed{our} experiments is available at \\
\href{https://github.com/takeuchi-lab/Learning-Interpretable-Metric-between-Graphs}{\tt https://github.com/takeuchi-lab/Learning-Interpretable-Metric-between-Graphs}. 

\section{Related Work}
\label{sec:related-work}

Kernel-based approaches have been widely studied for graph data analysis, and they can provide a metric of graph data in a reproducing kernel Hilbert space.
In particular, subgraph-based graph kernels are closely related to our study.
The graphlet kernel \citep{shervashidze2009efficient} creates a kernel through small subgraphs with only about 3--5 vertices, \ed{which are} called \red{graphlets}.
The neighborhood subgraph pairwise distance kernel \citep{costa2010fast} selects pairs of subgraphs from a graph and counts the number of \ed{pairs identical to those in} another graph.
The subgraph matching kernel \citep{kriege2012subgraph} identifies common subgraphs based on cliques in the product graph of two graphs.
%
%
%
\red{
The feature space created by these subgraph-based kernels is easy to interpret.
However, because the above approaches are unsupervised, it is fundamentally impossible to eliminate subgraphs that are unnecessary for a specific target classification task.
Therefore, for example, to create the entire kernel matrix of training data, all the candidate subgraphs in the data must be enumerated once, which becomes intractable even for small-sized subgraphs.
In contrast, we consider dynamically ``pruning'' unnecessary subgraphs through a supervised formulation of metric learning.
As we will demonstrate in our later experimental results, this significantly reduces the enumeration cost, allowing our proposed algorithm to deal with the larger size of subgraphs than the simple subgraph based kernels.
}
%
%

There are many other kernels including the shortest path \citep{borgwardt2005shortest}-, random walk \citep{vishwanathan2010graph, sugiyama2015halting, zhang2018retgk}-, and spectrum\ed{-based} \citep{kondor2008skew, kondor2009graphlet, kondor2016multiscale, verma2017hunt} approaches.
The Weisfeiler--Lehman (WL) kernel \citep{shervashidze2009fast,shervashidze2011weisfeiler}, which is based on the graph isomorphism test, is a popular and empirically successful kernel that has been employed in many studies \citep{yanardag2015deep,niepert2016learning,narayanan2017graph2vec,zhang2018end}.
\ed{Again, all such} approaches are unsupervised, and it is difficult to interpret results from the perspective of \ed{sub-structures} of a graph.
Although several kernels deal with continuous attributes on vertices \citep{feragen2013scalable,orsini2015graph,su2016fast,morris2016faster}, we only focus on the cases where vertex-labels are discrete \ed{due to the associated} interpretability.

\ed{Because} obtaining a good metric is an essential \ed{task} in data analysis, metric learning has been extensively studied \ed{to date}, as reviewed in \citep{li2018survey}.
However, due to \ed{its} computational difficulty, metric learning for graph data has not been widely studied.
A few studies \ed{have} considered the \emph{edit distance} approaches.
%
For example, \citet{bellet2012good} \ed{presented} a method \ed{for} learning a similarity function through an edit distance in a supervised manner.
%
Another approach \ed{probabilistically formulates} the editing process of the graph and estimates the parameters using labeled data \citet{neuhaus2007automatic}.
%
However, these approaches cannot provide any clear interpretation of the resulting metric in \ed{term} of the \red{subgraphs}.

\ed{Likewise,} the deep neural network (DNN) \ed{is} a standard approach to graph data analysis.
The deep graph kernel \citep{yanardag2015deep} incorporates neural language modeling, where decomposed sub-structures of a graph are regarded as \red{sentences}.
The PATCHY-SAN \citep{niepert2016learning} and DGCNN 
\red{\citep{zhang2018end}}
convert \red{a} graph to a tensor by using the WL-Kernel and convolute it. 
Several other studies also have combined popular convolution techniques with graph data \citep{tixier2018graph, atwood2016diffusion, simonovsky2017dynamic}.
These approaches are supervised, but the interpretability of these \ed{DNNs} is obviously \ed{relatively} low. 
\emph{Attention} enhances \ed{the} interpretability of deep learning, \ed{but} extracting important subgraphs is difficult because attention \ed{algorithms for graphs} \citep{lee2018graph} only provides the significance of vertex transition on a graph.
%
%
Another related DNN approach \ed{is} representation learning.
For example, sub2vec \citep{adhikari2018sub2vec} and graph2vec \citep{narayanan2017graph2vec} can embed graph data into a continuous space, but they are unsupervised, and it is difficult to extract substructures that characterize different classes. 
There are other fingerprint learning methods for graphs by neural networks \citep[e.g.][]{duvenaud2015convolutional} where the contribution from each node can be evaluated for each dimension of the fingerprint. 
%
Although it is possible to highlight sub-structures for the given input graph, this does not produce important common subgraphs for prediction.

Supervised pattern mining \citep{cheng2008direct,novak2009supervised,thoma2010discriminative} can be used for identifying important subgraphs by enumerating patterns with some discriminative score.
However, these approaches usually 1) employ \ed{a} greedy strategy to add a pattern \ed{for} which global optimality cannot be guaranteed, and 2) do not optimize a metric or representation.
%
\red{
A few other studies \citep{saigo2009gboost,nakagawa2016safe} have considered optimizing a linear model on the subgraph features with the LASSO penalty using graph mining.
A common idea of these two methods is to traverse a graph mining tree with pruning strategies derived from optimality conditions.
\citet{saigo2009gboost} employed a {boosting-based} approach, which adds a subgraph that violates the optimality condition most severely at every iteration.
It was shown that the maximum {violation} condition can be efficiently identified by pruning the tree without losing the final solution optimality. 
\citet{nakagawa2016safe} derived a pruning criterion by extending \emph{safe screening} \citep{ghaoui2010safe}, which can safely eliminate unnecessary features before solving the optimization problem.
This approach can {also} avoid enumerating the entire tree while guaranteeing the optimality, and {its} efficiency compared with the {boosting-based} approach was {demonstrated} empirically, mainly because it requires {much fewer tree traversals}. 
Further, \citet{morvan2018whinter} proposed a similar pruning extension of \emph{working set selection} for optimizing a higher-order interaction model.
Although this paper {was} not for the graph data, the technique is applicable to the same subgraph-based linear model as {in} \citep{saigo2009gboost} and \citep{nakagawa2016safe}.
Working set selection is a heuristic feature subset selection strategy that has been widely used in machine learning algorithms, such as support vector machines \citep[e.g.,][]{hsu2002simple}.
Unlike safe screening, this heuristic selection may eliminate necessary features in the middle of the optimization, but the optimality of the final solution can be guaranteed by iterating subset selection repeatedly until the solution converges.
However, these methods can only optimize a linear prediction model.
In this study, we focus on \emph{metric learning} of graphs.
Therefore, unlike the above mentioned pruning based learning methods, our aim is to learn a ``distance function''.
In metric learning, a distance function is typically learned from a loss function defined over a relative relation between samples (usually, pairs or triplets), by which a discriminative feature space that is generally effective for subsequent tasks, such as classification and similarity-based retrieval, is obtained.
Inspired by \citep{nakagawa2016safe} and \citep{morvan2018whinter}, we derive screening and pruning rules for this setting , and further, we combine them to develop an efficient algorithm.
}

\section{Formulation of Interpretable Graph Metric Learning}
\label{sec:learning}

\subsection{\red{Optimization problem}}



Suppose that the training dataset $\{(G_i,y_i)\}_{i \in [n]}$ consists of $n$ pairs of a graph $G_i$ and a class label $y_i$, where $[n] \coloneqq \{1, \ldots, n\}$.
%
Let $\cG$ be \ed{the} set of all \red{connected subgraphs} of $\{ G_i \}_{i \in [n]}$. 
In each graph, vertices and edges can be labeled.
%
If $H \in \cG$ is \red{a connected subgraph} of $G \in \cG$, we write $H \sqsubseteq G$.
%
%
Further, let $\#(H \sqsubseteq G)$ be the frequency of the subgraph $H$ in $G$.
%
Note that we adopt \ed{a} definition of frequency \ed{that} does not allow any vertices or edges among the counted subgraphs \ed{to overlap}.
%
As a representation of a graph $G$, we consider the following subgraph-based feature representation:
\[
 \phi_H(G) = g \bigl( \#(H\sqsubseteq G) \bigr), 
 \text{ for } H \in \cG,
 \addtag \label{eq:graph_feature} 
\]
where $g$ is some monotonically non-decreasing and non-negative function, such as the identity function $g(x) = x$ or indicator function $g(x) = 1_{x>0}$, which takes \ed{the value} $1$ if $x > 0$, and \ed{$0$} otherwise.
%
%
It is widely known that subgraph-based \ed{features can effectively represent} graphs.
%
For example, $g(x) = x$ allows all non-isomorphic graphs \ed{to be distinguished}. 
A similar idea was shown in \citep{gartner2003graph} \ed{for a frequency that} allows overlaps. 
%
%
However, this feature space is practically infeasible because the possible number of subgraphs is prohibitively large.

We focus on how to measure the distance between two graphs, which is essential for a variety of machine learning problems.
We consider the following weighted squared distance between two graphs:
\[
 d_{\*m}(G,G^\prime) \coloneqq
 \sum_{H \in \cG}
 m_{i(H)}
 \left(
 \phi_H(G) - \phi_H(G^\prime) 
 \right)^2, 
\]
where $i(H)$ is the index of the subgraph $H$ for a weight parameter $m_{i(H)} \geq 0$.
To obtain an effective and computable distance metric, we adaptively estimate $m_{i(H)}$ such that only a small number of important subgraphs have non-zero $m_{i(H)}$ \ed{values}.

%
Let $\*x_i \in \RR^p$ be the feature vector defined by concatenating $\phi_H(G_i)$ for all $H \in \cG$ included in the training dataset.
Then, we \ed{have}
\[
 d_{\bm m}(\bm x_i, \bm x_j)
 = 
 (\bm x_i-\bm x_j)^\top \mathrm{diag}(\bm m) (\bm x_i-\bm x_j)
 =\bm m^\top \bm c_{ij}, 
\]
where $\bm m\in\mathbb{R}_+^p$ is a vector of $m_{i(H)}$, and $\bm c_{ij} \in \RR^p$ is defined as $\bm c_{ij}\coloneqq(\bm x_i-\bm x_j)\circ (\bm x_i-\bm x_j)$ \ed{with} the element-wise product $\circ$.

Let 
$\mathcal{S}_i \subseteq [n]$ 
and
$\mathcal{D}_i \subseteq [n]$ 
be the subsets of indices that are in the same and different classes to $\*x_i$, respectively.
%
For \ed{each} of these sets, we select \ed{the} $K$ most similar inputs to $\*x_i$ by using some default metric, \red{such as the graph kernel (further details are presented in Section~\ref{ssec:SandD})}.
As a loss function for $\*x_i$, we consider
\begin{equation}
	\label{eq:loss}
	\ell_i(\*m ; L, U) 
	\coloneqq
	 \sum_{l\in \mathcal{D}_i}\ell_L(\bm m^\top \bm c_{il})+\sum_{j\in \mathcal{S}_i}\ell_{-U}(-\bm m^\top \bm c_{ij}), 
\end{equation}
where $L, U \in \RR_+$ are constant parameters satisfying $U \leq L$, and
$\ell_t(x) = [t-x]_+^2$ 
is the standard squared hinge loss function with threshold $t \in \RR$.
This loss function is a variant of the pairwise loss functions used in metric learning \citep{davis2007information}.
The first term in the loss function yields a penalty if $\*x_i$ and $\*x_l$ are closer than $L$ for $l \in \cD_i$, and the second term yields a penalty if $\*x_i$ and $\*x_j$ are more distant than $U$ for $j \in \cS_i$.

Let
$R(\bm m)=\|\bm m\|_1+\frac{\eta}{2}\|\bm m\|_2^2=\bm m^\top \bm1+\frac{\eta}{2}\|\bm m\|_2^2$
be an elastic-net type \ed{sparsity-inducing} penalty, where $\eta\ge0$ is a non-negative parameter.
We define our proposed IGML (\emph{interpretable graph metric learning}) as the following regularized loss minimization problem:
\[
 \begin{split}	 
  \min_{\bm m\ge\bm 0}
  P_{\lambda}(\bm m) 
  \coloneqq &
  \sum_{i\in[n]}
  \ell_i(\*m ; L, U) 
  + \lambda R(\bm m),
 \end{split} 
 \addtag \label{eq:primal}
\]
where $\lambda > 0$ is the regularization parameter.
The solution of this problem can provide not only a discriminative metric but also insight into important subgraphs because the sparse penalty is expected to select only a small number of non-zero parameters.
%
%


Let $\*\alpha \in \RR^{2nK}_+$ be the vector of dual variables where $\alpha_{il}$ and $\alpha_{ij}$ for $i \in [n], l \in \cD_i$, and $j \in \cS_i$ are concatenated.
%
%
The dual problem of \eqref{eq:primal} is written as follows (see Appendix~\ref{app:dual} for derivation): 
\begin{equation}
	\label{eq:dual}
	\max_{\bm \alpha\ge\bm0}
	D_\lambda(\bm\alpha)\coloneqq-\frac{1}{4}\|\bm\alpha\|_2^2+\bm t^\top\bm\alpha
	-\frac{\lambda\eta}{2}\|\bm m_{\lambda} (\bm\alpha)\|_2^2, 
\end{equation}
where
\[
 \bm m_\lambda(\bm\alpha)\coloneqq 
 \frac{1}{\lambda\eta}[\bm C\bm \alpha-\lambda\bm 1]_+, 
 \addtag \label{eq:malpha}
\]
$\bm t\coloneqq[L,\ldots,L,-U,\ldots,-U]^\top\in\mathbb{R}^{2nK}$
and
$\bm C\coloneqq[\ldots,\bm c_{il},\ldots,$ $-\bm c_{ij}, \ldots] \in \mathbb{R}^{p\times 2nK}$.
%
%
Then, from the optimality condition, we obtain the following relationship between the primal and dual variables: 
\[
  \alpha_{il} =-\ell'_L(\bm m^\top\bm c_{il}), ~
  \alpha_{ij} =-\ell'_{-U}(-\bm m^\top\bm c_{ij}), 
  \addtag \label{eq:primal2dual}
\]
where $\ell_t'(x)=-2[t-x]_+$ is the derivative of $\ell_t$.
When the regularization parameter $\lambda$ is larger than certain $\lambda_{\max}$, the optimal solution is $\bm m = \bm 0$.
Then, the optimal dual variables are $\alpha_{il}=-\ell_L'(0)=2L$ and $\alpha_{ij}=-\ell_{-U}'(0)=0$.
By substituting these equations into \eqref{eq:malpha}, we obtain $\lambda_{\max}$ as
\[
  \lambda_{\max} = \max_{k}\bm C_{k,:}\bm\alpha.
  \addtag \label{eq:lambdaMax}
\]


\subsection{\red{Selection of $\cS_i$ and $\cD_i$}}
\label{ssec:SandD}

\red{
For $K = |\cS_i| = |\cD_i|$, in {the} experiments {reported later}, we employed the small number $K = 10$ and used a graph kernel to select samples in $\cS_i$ and $\cD_i$.
%
Although we simply used {a} pre-determined kernel, selecting {the} kernel (or its parameter) through cross-validation beforehand is also possible.
Using only a small number of neighbors is a common setting in metric learning. 
For example, \citet{davis2007information}, which is a seminal work {on} the pairwise approach, only used $20c^2$ pairs in total, where $c$ is the number of classes.
%
{A} small $K$ setting {has} two aims.
%
First, {particularly} $\cS_i$, adding pairs {that are too far apart can be avoided under} this setting.
Even for a pair of samples with the same labels, enforcing such distant pairs to be close may cause over-fitting (e.g., when the sample is an outlier).
Second, a small $K$ reduces the computational cost.
Because the number of pairs is $O(n^2)$, adding all of them into the loss term requires a large computational cost.
%
In fact, these two issues are not only for the pairwise formulation but also for other relative loss functions such as the standard triplet loss, for which there exist $O(n^3)$ triplets.
One potential difficulty in selecting $\cD_i$ and $\cS_i$ is the discrepancy between the initial and the optimal metric.
The loss function is defined through $\cD_i$ and $\cS_i$, which are selected based on the neighbors in the initial metric, but the optimization of the metric may change the nearest neighbors of each sample.
%
A possible remedy for this problem is to adaptively change $\cD_i$ and $\cS_i$ in accordance with the updated metric \citep{takeuchi2011target}, though the resulting optimality of this approach is still not known.
%
To the best of our knowledge, this is still an open problem in metric learning, which we consider beyond the scope of this paper.
In the experiments (Section~\ref{sec:experiments}), we show that a nearest-neighbor classifier in the learned metric with this heuristics selection of $\cD_i$ and $\cS_i$ shows better or comparable performance to standard graph classification methods, such as a graph neural network.
}

\section{Creating \ed{a} Tractable Sub-problem}
\label{sec:safe-and-working}

Because the problems \ed{of} \eqref{eq:primal} and \eqref{eq:dual} are convex, the local solution is equivalent to the global optimal.
However, na{\"i}vely solving these problems is computationally intractable because of the high dimensionality of $\*m$.
In this section, we introduce several useful rules for restricting candidate subgraphs while maintaining the optimality of the final solution. 
%
Note that the proofs for all the \ed{lemmas} and theorems are provided in the appendix.


To make the optimization problem tractable, we work with only a small subset of features during the optimization process.
Let 
$\cF \subseteq [p]$
be a subset of features. 
%
By fixing $m_i = 0$ for $i \notin \cF$,
we define sub-problems of the original primal $P_{\lambda}$ and dual $D_{\lambda}$ problems as follows:
\[
 \min_{{\bm m}_\cF\ge\bm 0}
 P_{\lambda}^\cF({\bm m}_\cF)
 \coloneqq
 \sum_{i\in[n]}[\sum_{l\in \mathcal{D}_i}\ell_L({\bm m}_\cF^\top {\bm c_{il}}_\cF)
 +\sum_{j\in \mathcal{S}_i}\ell_{-U}(-{\bm m}_\cF^\top {\bm c_{ij}}_\cF)] 
 +\lambda R({\bm m}_\cF),
 \addtag \label{eq:primal-sub}
\]
\[
 \max_{\bm \alpha\ge\bm0}
 D_\lambda^\cF(\bm\alpha)
 \coloneqq-\frac{1}{4}\|\bm\alpha\|_2^2+\bm t^\top\bm\alpha
 -\frac{\lambda\eta}{2}\|\bm m_{\lambda} (\bm\alpha)_\cF\|_2^2,
 \addtag \label{eq:dual-sub}
\]
where $\*m_{\cF}$, $\*c_{ij\cF}$, and $\*m_{\lambda} (\bm\alpha)_\cF$ are sub-vectors specified by $\cF$.
%
If the size of $\cF$ is moderate, these sub-problems are \ed{significantly} computationally easier to solve than the original problems.

We introduce several criteria that determine whether the feature $k$ should be included in $\cF$ \ed{using the} techniques of \emph{safe screening} \citep{ghaoui2010safe} and \emph{working set selection} \citep{fan2008liblinear}.
A general form of our criteria can be written as
\[
 \bm C_{k,:}\bm q+r\|\bm C_{k,\red{:}}\|_2 \leq T,
 \addtag \label{eq:general-rule}
\]
where $\bm q\in\mathbb{R}_+^{2nK}$, $r\ge0$, and $T \in \RR$ are constants that assume different values \ed{for each} criterion.
If this inequality holds for $k$, we exclude the $k$-th feature from $\cF$. 
An important property is that although our algorithm only solves these small sub-problems, we can guarantee the optimality of the final solution, as shown later. 

\begin{figure}[t]
 \centering
 \includegraphics[clip,width=0.65\linewidth]{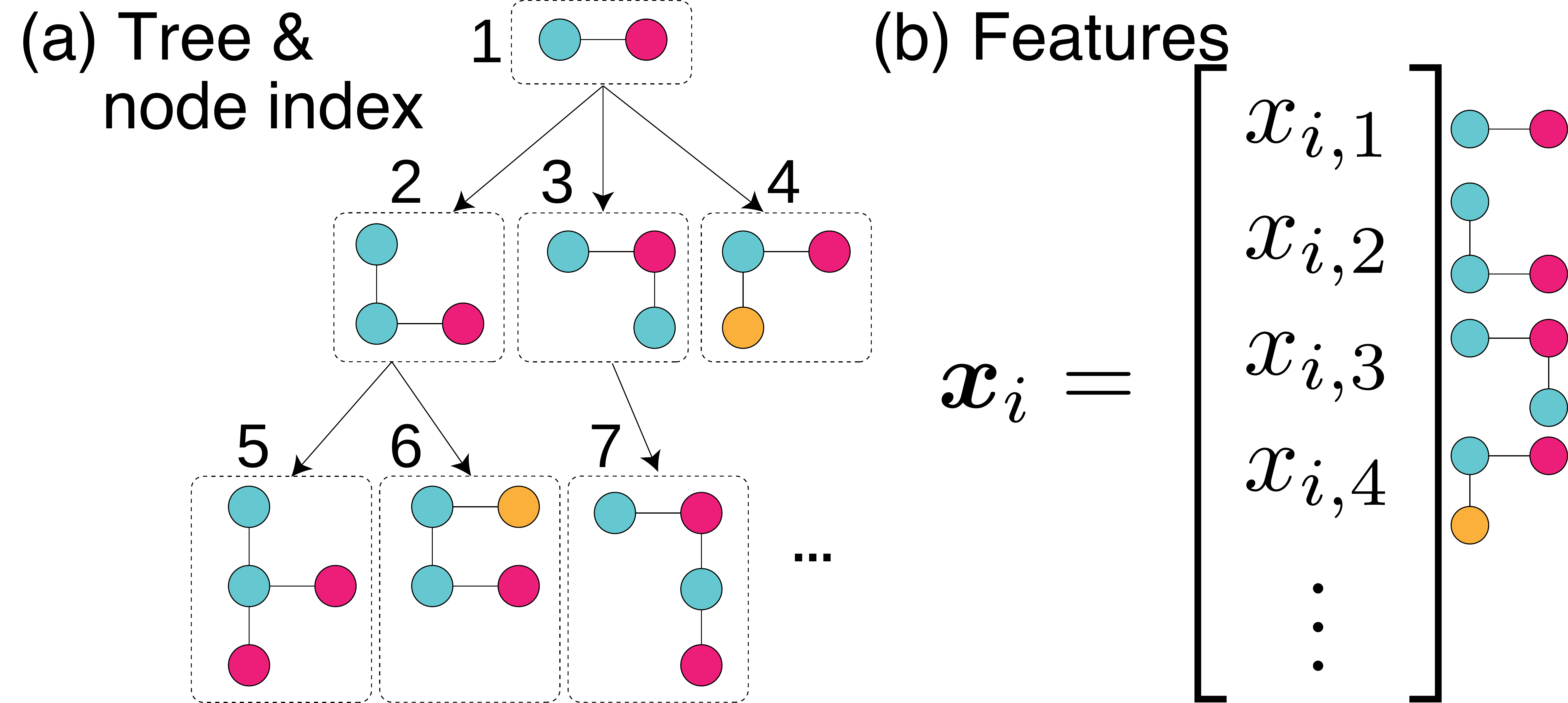}  
 \caption{Schematic illustration of \ed{a} tree and \ed{features}.}
 \label{fig:tree}
\end{figure}

However, selecting $\cF$ itself is computationally expensive \ed{because} the evaluation of \eqref{eq:general-rule} requires $O(n)$ computations for each $k$.
%
%
Thus, we exploit a tree structure of graphs for determining $\cF$.
\figurename~\ref{fig:tree} shows an example of \ed{such a} tree, which can be constructed by \ed{a} graph mining algorithm, such as gSpan \citep{yan2002gspan}.
Suppose that the $k$-th node corresponds to the $k$-th dimension of $\*x$ (\ed{note} that the node index \ed{here} is not the order of the visit).
%
If a graph corresponding to the $k$-th node is a subgraph of the $k'$-th node, the node $k'$ is a descendant of $k$, which is denoted as $k' \supseteq k$.
Then, the following monotonic relation is immediately derived from the monotonicity of $\phi_H$:
\[
 x_{i,k'} \leq x_{i,k} \text{ if } k' \supseteq k.
 \addtag \label{eq:monotonicity}
\]
%
%
\red{
Because any parent node is a subgraph of its children in the gSpan tree \figurename~\ref{fig:tree}, the non-overlapped frequency $\#(H \sqsubseteq G)$ of subgraph $H$ in $G$ is monotonically non-increasing while descending the tree node.
Then, the condition (\ref{eq:monotonicity}) is obviously satisfied because for a sequence of $H \sqsubseteq H' \sqsubseteq H'' \sqsubseteq \cdots$ in the descending path of the tree, $x_{i,k(H)} = \phi_H(G_i) = g(\#(H \sqsubseteq G))$ is monotonically non-increasing, where $x_{i,k(H)}$ is a feature corresponding to $H$ in $G_i$.
}
%
Based on this property, the following lemma enables us to prune a node during the tree traversal.
\begin{lemma}
 \label{lmm:pruning}
 Let
\begin{align} 
 \mathrm{Prune}(k | \bm q, r) \!\coloneqq\! 
 \sum_{i\in[n]}\sum_{l\in \mathcal{D}_i}q_{il}\max\{x_{i,k},x_{l,k}\}^2 
  +r\sqrt{\sum_{i\in[n]}[\sum_{l\in\mathcal{D}_i}\max\{x_{i,k},x_{l,k}\}^4+\sum_{j\in\mathcal{S}_i}\max\{x_{i,k},x_{j,k}\}^4]}
 \label{eq:pruning}
\end{align}
 be a pruning criterion.
 %
 %
 Then, if the inequality
 \[
 \mathrm{Prune}(k | \bm q, r) \leq T
 \addtag \label{eq:general-pruning}
 \]
 holds, for any descendant node $k' \supseteq k$, the following inequality holds: 
 \[
 \bm C_{k',:}\bm q+r\|\bm C_{k',;}\|_2 \le T, 
 \]
 where $\bm q\in\mathbb{R}_+^{2nK}$ and $r\ge0$ are \ed{an} arbitrary constant vector and scalar \ed{variable, respectively}.
\end{lemma}
\noindent
This lemma indicates that if the condition \eqref{eq:general-pruning} is satisfied, we can \ed{say} that none of the descendant nodes are included in $\cF$. 
%
Assuming that the indicator function $g(x) = 1_{x>0}$ is used in \eqref{eq:graph_feature}, a tighter bound can be obtained \ed{through the following lemma}.
\begin{lemma}
 \label{lmm:pruning-binary}
 If $g(x) = 1_{x>0}$ is set in \eqref{eq:graph_feature}, the pruning criterion \eqref{eq:pruning} can be replaced with
 {\small
 \begin{multline*}
  \mathrm{Prune}(k | \bm q, r)\!\coloneqq\!\sum_{i\in[n]}\max\{\sum_{l\in \mathcal{D}_i}q_{il}x_{l,k}, x_{i,k}[\sum_{l\in \mathcal{D}_i}q_{il}-\sum_{j\in \mathcal{S}_i}q_{ij}(1-x_{j,k})]\}\\
  +r\sqrt{\sum_{i\in[n]}[\sum_{l\in\mathcal{D}_i}\max\{x_{i,k},x_{l,k}\}+\sum_{j\in\mathcal{S}_i}\max\{x_{i,k},x_{j,k}\}]}.
 \end{multline*}}
\end{lemma}
\noindent
By comparing the first terms of \ed{Lemmas~\ref{lmm:pruning} and \ref{lmm:pruning-binary}}, we see that \ed{Lemma}~\ref{lmm:pruning-binary} is tighter when $g(x) = 1_{x>0}$ as follows:
{\small\begin{align*}
\sum_{i\in[n]}\max\{\sum_{l\in \mathcal{D}_i}q_{il}x_{l,k}, x_{i,k}[\sum_{l\in \mathcal{D}_i}q_{il}-\sum_{j\in \mathcal{S}_i}q_{ij}(1-x_{j,k})]\}
&\le 
\sum_{i\in[n]}\max\{\sum_{l\in \mathcal{D}_i}q_{il}x_{l,k}, x_{i,k}\sum_{l\in \mathcal{D}_i}q_{il}\}\\
&=\sum_{i\in[n]}\max\{\sum_{l\in \mathcal{D}_i}q_{il}x_{l,k}, \sum_{l\in \mathcal{D}_i}q_{il}x_{i,k}\}\\
&\le \sum_{i\in[n]}\sum_{l\in \mathcal{D}_i}\max\{q_{il}x_{l,k}, q_{il}x_{i,k}\}\\
&=\sum_{i\in[n]}\sum_{l\in \mathcal{D}_i}q_{il}\max\{x_{l,k}, x_{i,k}\}.
\end{align*}}
%

\begin{figure*}[tbp]
	\centering
	\includegraphics[width=\linewidth]{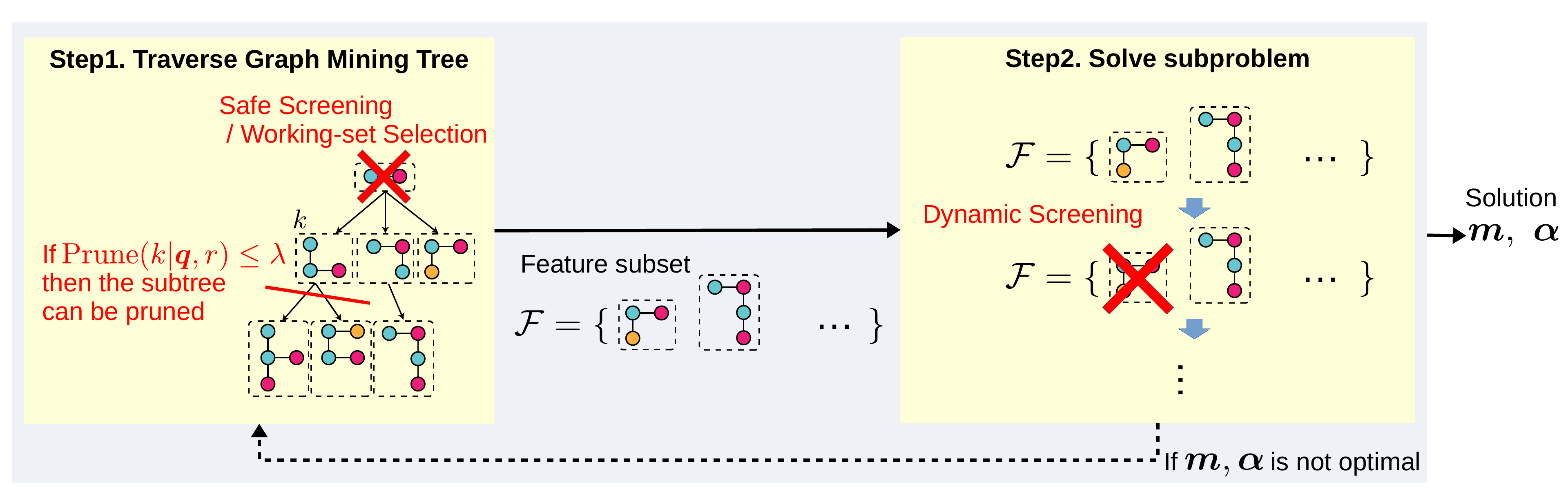}
	\caption{Schematic illustration of \ed{the} optimization algorithm for IGML. }
	\label{fig:opt}
\end{figure*}

A schematic illustration of \ed{the} optimization algorithm for IGML is shown in \figurename~\ref{fig:opt} (for \ed{further} details, see Section~\ref{sec:computations}). 
To generate a subset of features $\mathcal{F}$, we first traverse the graph mining tree during which the safe screening/working set selection procedure and their pruning extensions are performed (Step1). 
Next, we solve the \ed{sub-problem} \eqref{eq:primal-sub} with the generated $\mathcal{F}$ using a standard gradient-based algorithm (Step2). 
Safe screening is also performed during the optimization iteration in \ed{Step2}, which is referred to as \emph{dynamic screening}. 
This further reduces the size of $\mathcal{F}$. 

\red{ 
Before moving onto detailed formulations, we summarize our rules to determine $\cF$ in \tablename~\ref{tab:rules}.
The columns represent the different approaches to evaluating the necessity of features, i.e., safe and working set approaches.
%
For the safe approaches, there are further `single $\lambda$' (described in Section~\ref{sssec:safe}) and `range of $\lambda$' (described in Section~\ref{sssec:range-based}) approaches.
The single $\lambda$ approach considers safe rules for a specific $\lambda$, while the range of $\lambda$ approach considers safe rules that can eliminate features for a range of $\lambda$ (not just a specific value).
Both the single and range approaches are based on the bounds of the region in which the optimal solution exists, for which details are given in Section~\ref{sssec:bound}.
%
The rows of \tablename~\ref{tab:rules} indicate the variation of rules to remove one specific feature and rules to prune all features in a subtree.
}

\begin{table}[t]
 \centering
 \caption{\red{Strategies to determine $\cF$.}}
 \label{tab:rules}
\red{
 \begin{tabular}{|l||c|c|c|}
  & \multicolumn{2}{|c|}{Safe approaches}  & Working set \\
  & \multicolumn{2}{|c|}{based on the bounds in Sec.~\ref{sssec:bound}} & approaches \\
  & Single $\lambda$ & Range of     $\lambda$ & \\ 
  & (Sec.~\ref{sssec:safe})     & (Sec.~\ref{sssec:range-based}) & (Sec.~\ref{sssec:ws-wp})
  \\ \hline \hline
  For removing & Safe screening (SS)   & Range-based safe screening & Working set selection \\
  a single feature & rule & (RSS) & (WS)
  \\
  \hline
  For pruning & Safe pruning (SP) & Range-based safe pruning & Working set pruning \\ 
  a subtree & rule &  (RSP) & (WP) \\ 
  \hline
 \end{tabular}
}
\end{table}

\subsection{Safe Screening}

Safe screening \citep{ghaoui2010safe} was first proposed to identify unnecessary features in LASSO-type problems.
Typically, this approach considers a bounded region of dual variables \ed{in which} the optimal solution must exist.
Then, we can eliminate dual inequality constraints \ed{that} are never violated \ed{given that} the solution exists in that region.
The well-known Karush-Kuhn-Tucker (KKT) conditions show that this is equivalent to the elimination of primal variables \ed{that take value} $0$ at the optimal solution.
In Section~\ref{sssec:bound}, we first derive a spherical bound \ed{for} our optimal solution, and \ed{then} in Section~\ref{sssec:safe}, a rule for safe screening is shown.
%
Section~\ref{sssec:range-based} \ed{extends} rules that are specifically useful for the regularization path calculation. 

\subsubsection{Sphere Bound for Optimal Solution}
\label{sssec:bound}

%
The following theorem provides a hyper-sphere containing the optimal dual variable $\*\alpha^\star$.
\begin{theorem}[DGB] \label{thm:DGB}
 For any pair of $\bm m\ge\bm 0$ and $\bm \alpha\ge\bm 0$, the optimal dual variable $\bm \alpha^\star$ must satisfy
 \[
 \|\bm \alpha-\bm \alpha^\star\|_2^2\le 4(P_\lambda(\bm m)-D_\lambda(\bm \alpha)).
 \]
\end{theorem}
\noindent
This bound is called \ed{the} \emph{duality gap bound} (DGB), and the parameters $\*m$ and $\*\alpha$ used to construct the bound are referred to as the \emph{reference solution}.
\red{
This inequality reveals that the optimal $\*\alpha^{\star}$ should be in the inside of the sphere whose center is the reference solution $\*\alpha$ and radius is $2 \sqrt{P_\lambda(\bm m)-D_\lambda(\bm \alpha)}$, i.e., twice the square root of the duality gap.
Therefore, if the quality of the reference solution $\*m$ and $\*\alpha$ is better, a tighter bound can be obtained.
When the duality gap is zero, meaning that $\*m$ and $\*\alpha$ are optimal, the radius is shrunk to zero.
}

If the optimal solution for $\lambda_0$ is available as a reference solution to construct the bound for $\lambda_1$, the following bound, called \emph{regularization path bound} (RPB), can be obtained.
\begin{theorem}[RPB] \label{thm:RPB}
Let 
$\bm \alpha_0^\star$
be the optimal solution for $\lambda_0$ and 
$\bm \alpha_1^\star$
be the optimal solution for $\lambda_1$.
\ed{Then,}
 \[
 \left\|\bm \alpha_1^\star-\frac{\lambda_0+\lambda_1}{2\lambda_0}\bm \alpha_0^\star\right\|_2^2
 \le \left\|\frac{\lambda_0-\lambda_1}{2\lambda_0}\bm \alpha_0^\star\right\|_2^2.
 \]
\end{theorem}
\noindent
\red{
This inequality indicates that the optimal dual solution for $\lambda_1$ ($\*\alpha_1^{\star}$) should be in the sphere whose center is 
$\frac{\lambda_0+\lambda_1}{2\lambda_0}\bm \alpha_0^\star$ and radius is 
$\left\|\frac{\lambda_0-\lambda_1}{2\lambda_0}\bm \alpha_0^\star\right\|_2$.
}
%
However, RPB requires the exact solution, which is difficult \ed{to obtain} in practice \ed{due to} numerical errors. 
%
\ed{The} \emph{relaxed RPB} (RRPB) extends RPB \ed{to} incorporate the approximate solution as a reference solution.
\begin{theorem}[RRPB] \label{thm:RRPB}
 Assuming that $\bm\alpha_0$ satisfies $\|\bm \alpha_0-\bm \alpha_0^\star\|_2\le\epsilon$,
 \ed{the} optimal solution $\bm \alpha_1^\star$ for $\lambda_1$ must satisfy
 \[
 \left\|\bm \alpha_1^\star-\frac{\lambda_0+\lambda_1}{2\lambda_0}\bm \alpha_0\right\|_2
 \le \left\|\frac{\lambda_0-\lambda_1}{2\lambda_0}\bm \alpha_0\right\|_2+\Bigl(\frac{\lambda_0+\lambda_1}{2\lambda_0}+\frac{|\lambda_0-\lambda_1|}{2\lambda_0}\Bigr)\epsilon.
 \]
\end{theorem}
\noindent
\red{
In Theorem~\ref{thm:DGB}, the reference $\*\alpha_0$ is only assumed to be close to $\*\alpha_0^\star$ within the radius $\epsilon$ instead of assuming that $\*\alpha_0^\star$ is available.
}
%
For example, $\epsilon$ can be obtained using the DGB (Theorem~\ref{thm:DGB}).

Similar bounds to \ed{those derived} here were previously considered for the triplet screening of metric learning on usual numerical data \citep{yoshida2018safe,yoshida2019safe}.
Here, we extend a similar idea to derive subgraph screening.

\subsubsection{Safe Screening and Safe Pruning Rules}
\label{sssec:safe}

Theorem~\ref{thm:DGB} and \ref{thm:RRPB} identify the regions where \red{the} optimal solution exists \ed{using} a current feasible solution $\bm\alpha$.
Further, from \eqref{eq:malpha}, when $\bm C_{k,:}\bm\alpha^\star\le\lambda$, we \ed{have} $m_k^\star=0$.
This indicates that 
\[
 \max_{\bm \alpha\in\mathcal{B}}\bm C_{k,:}\bm\alpha \le \lambda\Rightarrow m_k^\star=0,
 \addtag \label{eq:screeningRuleRaw}
\]
where $\cal B$ is a region containing the optimal solution $\bm \alpha^\star$, i.e., $\bm \alpha^\star\in\mathcal{B}$.
\red{
As we derived in Section~\ref{sssec:bound}, the sphere-shaped $\cal B$ can be constructed using feasible primal and dual solutions.
}
By solving this maximization problem, we obtain the following safe screening \ed{(SS)} rule.
%
\begin{theorem}[SS Rule] 
 \label{thm:SS}
 If the optimal solution $\bm \alpha^\star$ exists in the bound $\mathcal{B}=\{\bm \alpha \mid \| \bm\alpha -\bm q \|_2^2\le r^2\}$, the following rule holds
 \[  
  \bm C_{k,:}\bm q+r\|\bm C_{k,:}\|_2 \le \lambda \Rightarrow m_k^\star=0.
 \addtag   \label{eq:screeningRule}
 \]
\end{theorem}
%
\noindent
Theorem~\ref{thm:SS} indicates that we can eliminate unnecessary features by evaluating the condition shown in \eqref{eq:screeningRule}.
\red{
Here, the theorem is written in a general form, and in practice, $\*q$ and $r$ can be defined by the center and a radius of one of the sphere bounds, respectively. 
}
An important property of this rule is that it guarantees optimality, meaning that the sub-problems \eqref{eq:primal-sub} and \eqref{eq:dual-sub} have the exact same optimal solution to the original problem if $\cF$ is defined through this rule.
%
However, it is still necessary to evaluate the rule for all $p$ features, which is currently intractable.
To avoid this problem, we derive a pruning strategy on the graph mining tree, which we call \ed{the} safe pruning \ed{(SP)} rule.
\begin{theorem}[SP Rule] \label{thm:SP}
 If the optimal solution $\bm \alpha^\star$ is in the bound $\mathcal{B}=\{\bm \alpha \mid \| \bm\alpha -\bm q \|_2^2\le r^2, \bm q\ge\bm0\}$, the following rule holds 
  \[
  \mathrm{Prune}(k|\bm q, r)\le\lambda \Rightarrow m_{k'}^\star=0.
   \text{ for } \forall k' \supseteq k.
   \addtag \label{eq:pruningRule}
  \]
\end{theorem}
\noindent
This theorem is a direct consequence of \ed{Lemma}~\ref{lmm:pruning}.
If this condition holds for a node $k$ during the tree \ed{traversal}, a subtree below that node can be pruned.
This means that we can safely eliminate unnecessary subgraphs even without enumerating them.
\red{In this theorem, note that $\cB$ has an additional non-negative constraint $\*q \geq \*0$, but this is satisfied by all the bounds in Section~\ref{sssec:bound} because of the non-negative constraint in the dual problem.}

\subsubsection{Range-based Safe Screening \& \ed{Safe Pruning}}
\label{sssec:range-based}

\ed{The} SS and SP \ed{rules apply to} a fixed $\lambda$.
%
The range-based extension identifies an interval of $\lambda$ \ed{for which} the satisfaction of SS/SP is guaranteed.
This is particularly useful for the \emph{path-wise optimization} or \emph{regularization path} calculation, where the problem \ed{must be solved} with a sequence of $\lambda$.
%
We assume that the sequence is sorted in descending order, as optimization algorithms typically start from the trivial solution $\*m = \*0$.
Let 
$\lambda=\lambda_1\le\lambda_0$.
By combining RRPB with the rule \eqref{eq:screeningRule}, we obtain the following theorem.
\begin{theorem}[Range-based Safe Screening (RSS)]
 \label{thm:RSS}
 For any $k$, the following rule holds
 \[  
  \lambda_a\le \lambda\le\lambda_0 \Rightarrow m_k^\star=0,
  \addtag \label{eq:RSS}
 \]
 where
 \[
  \lambda_a \coloneqq \frac{\lambda_0(2\epsilon\|\bm C_{k,:}\|_2
  +\|\bm\alpha_0\|_2\|\bm C_{k,:}\|_2
  +\bm C_{k,:}\bm\alpha_0)}{2\lambda_0+\|\bm\alpha_0\|_2\|\bm C_{k,:}\|_2
  -\bm C_{k,:}\bm\alpha_0}
  .
 \]
\end{theorem}
\noindent
\red{
This rule indicates that we can safely ignore $m_k$ 
for
$\lambda \in [\lambda_a, \lambda_0]$, while if $\lambda_a > \lambda_0$, the weight $m_k$ cannot be removed by this rule.
}
%
For SP, the range-based rule can also be derived from \eqref{eq:pruningRule}.
\begin{theorem}[Range-based Safe Pruning (RSP)]
 \label{thm:RSP}
 For any 
 $k' \supseteq k$,
 the following pruning rule holds:
 \[
 \lambda'_a \coloneqq
 \frac{\lambda_0(2\epsilon b+\|\bm\alpha_0\|_2b+a)}{2\lambda_0+\|\bm\alpha_0\|_2b-a}\le \lambda\le\lambda_0
 \Rightarrow
 m_{k'}^\star=0,
 \addtag \label{eq:RSP}
 \] 
 where
 {\small\begin{align*}
	a&\coloneqq
	\sum_{i\in[n]}\sum_{l\in \mathcal{D}_i}{\alpha_0}_{il}\max\{x_{l,k}, x_{i,k}\}^2,\\
	b&\coloneqq\sqrt{\sum_{i\in[n]}[\sum_{l\in\mathcal{D}_i}\max\{x_{i,k},x_{l,k}\}^4+\sum_{j\in\mathcal{S}_i}\max\{x_{i,k},x_{j,k}\}^4]}.
	\end{align*}}
\end{theorem}
\noindent
\red{This theorem indicates that, while $\lambda \in [\lambda_a',\lambda_0]$, we can safely remove the entire subtree of $k$.}
\ed{Analogously}, if the feature vector is generated from $g(x) = 1_{x>0}$ (i.e., binary), the following theorem holds.
\begin{theorem}[Range-Based Safe Pruning (RSP) for binary feature]
 \label{thm:RSP-bin}
Assuming $g(x) = 1_{x>0}$ in \eqref{eq:graph_feature}, $a$ and $b$ in theorem~\ref{thm:RSP} can be replaced with
 {\small\begin{align*}
 	 a&\coloneqq
 	 \sum_{i\in[n]}\max\{\sum_{l\in \mathcal{D}_i}{\alpha_0}_{il}x_{l,k}, x_{i,k}[\sum_{l\in \mathcal{D}_i}{\alpha_0}_{il}-\sum_{j\in \mathcal{S}_i}{\alpha_0}_{ij}(1-x_{j,k})]\},\\
 	 b&\coloneqq\sqrt{\sum_{i\in[n]}[\sum_{l\in\mathcal{D}_i}\max\{x_{i,k},x_{l,k}\}+\sum_{j\in\mathcal{S}_i}\max\{x_{i,k},x_{j,k}\}]}.
 \end{align*}}
\end{theorem}
\noindent
\ed{Because} these constants $a$ and $b$ are derived from the tighter bound in \ed{Lemma}~\ref{lmm:pruning-binary}, the obtained range becomes wider than the range in \ed{Theorem}~\ref{thm:RSP}.

\ed{Once} we calculate $\lambda_a$ and $\lambda'_a$ of \eqref{eq:RSS} and \eqref{eq:RSP} for some $\lambda$, they are stored at each node of the tree.
%
Subsequently, \ed{such} $\lambda_a$ and $\lambda'_a$ can be used for the next tree \ed{traversal} with different $\lambda'$.
%
If the \ed{conditions of} \eqref{eq:RSS} or \eqref{eq:RSP} \ed{are} satisfied, the node can be skipped (RSS) or pruned (RSP).
Otherwise, we update $\lambda_a$ and $\lambda'_a$ by using the current reference solution.


\subsection{Working Set Method}
\label{sssec:working}

Safe rules are strong rules in \ed{the} sense that they can completely remove features\ed{; thus, they are sometimes} too conservative to fully accelerate the optimization.
In contrast, the \emph{working set selection} is a widely accepted heuristic approach to selecting a subset of features.

\subsubsection{Working Set Selection \& Working Set Pruning}
\label{sssec:ws-wp}

\ed{The} working set (WS) method optimizes the problem with respect to \ed{only} selected working set features.
%
Then, if the optimality condition for the original problem is not satisfied, the working set is \ed{reselected} and the optimization on the new working set \ed{restarts}.
This process iterates until optimality on the original problem is achieved.

Besides the safe rules, we use the following WS selection criterion, which is obtained directly from the KKT conditions:
\begin{equation}\label{eq:workingset}
	\bm C_{k,:}\bm\alpha\le\lambda.
\end{equation}
If this inequality is satisfied, the $k$-th dimension is predicted as $m_k^\star=0$.
Hence, the working set is defined by
\[
 \mathcal{W}\coloneqq\{k\mid \bm C_{k,:}\bm\alpha>\lambda\}.
\]
%
Although $m^\star_i = 0$ for $i \notin \cW$ is not guaranteed, the final convergence of the procedure is \ed{guaranteed} by the following theorem.
\begin{theorem}[Convergence of WS]
 \label{thm:WS}
 Assume that there is a solver for the sub-problem \eqref{eq:primal-sub} (or equivalently \eqref{eq:dual-sub}) \ed{that} returns the optimal solution for given $\cF$.
 \ed{The} working set method, which iterates optimizating the sub-problem with $\cF = \cW$ and updating $\cW$ alternately, returns the optimal solution of the original problem \ed{in} finite steps.
\end{theorem}
\noindent
However, here again, the inequality \eqref{eq:workingset} needs to be evaluated for all features, which is computationally intractable.

The same pruning strategy as \ed{for} SS/SP can be incorporated into working set selection.
The criterion \eqref{eq:workingset} is also a special case of \eqref{eq:general-rule}, and
%
\ed{Lemma}~\ref{lmm:pruning} indicates that if the following inequality
\[
\mathrm{Prune}_{\rm WP}(k) \coloneqq \mathrm{Prune}(k|\bm \alpha, 0)\le\lambda,
\]
%
holds, then \ed{no} $k' \supseteq k$ is included in the working set.
We refer to this criterion as working set pruning (WP).

\subsubsection{Relation with Safe Rules}
\label{sssec:relation-with-safe-rules}

Note that for \ed{the} working set method, we may need to update $\cW$ multiple times, unlike \ed{in the} safe screening approaches, \ed{as shown by Theorem}~\ref{thm:WS}.
Instead, \ed{the} working set method can usually exclude a larger number of features compared with safe screening approaches. 
%
In fact, when the condition of the SS rule \eqref{eq:screeningRule} is satisfied,  
the WS criterion \eqref{eq:workingset} must likewise \ed{be} satisfied.
%
\ed{Because} all the spheres (DGB, RPB and RRPB) contain the reference solution $\*\alpha$, \ed{which is usually the current solution}, the inequality
\[
 \*C_{k,:} \*\alpha \leq \max_{\*\alpha' \in \cB} \*C_{k,:} \*\alpha'
 \addtag \label{SSandWS}
\]
holds, where $\cB$ is a sphere created by DGB, RPB or RRPB.
This indicates that when the SS rule excludes the $k$-th feature, the WS also excludes the $k$-th feature.
However, to guarantee convergence, WS needs to be fixed until the sub-problem \eqref{eq:primal-sub}--\eqref{eq:dual-sub} is solved (\ed{Theorem}~\ref{thm:WS}).
In contrast, the SS rule is applicable anytime during the optimization procedure without affecting the final optimality.
%
This enables us to apply the SS rule even to the sub-problem \eqref{eq:primal-sub}--\eqref{eq:dual-sub}, \ed{where} $\cF$ is defined by WS as shown in \ed{Step}~2 of \figurename~\ref{fig:opt} (dynamic screening).
%
%

For the pruning rules, we first confirm the following two properties: 
\begin{align*}
 {\rm Prune}(k|\*q,r) &\geq {\rm Prune}(k|\*q,0), \\
 {\rm Prune}(k| C \*q,0) &= C ~ {\rm Prune}(k| \*q,0),
\end{align*}
where $\*q \in \RR_+^{2 n K}$ is the center of the sphere, $r \geq 0$ is the radius, and $C \in \RR$ is a constant.
%
%
In the case of DGB, the center 
of the sphere 
is the reference solution $\*\alpha$ itself, i.e., $\*q = \*\alpha$.
Then, the following relation holds between the SP criterion ${\rm Prune}(k|\*q,r)$ and WP criterion ${\rm Prune}_{\rm WP}(k)$:
%
\begin{align*}
 {\rm Prune}(k|\*q,r) =
 {\rm Prune}(k|\*\alpha_0,r) 
 \geq
 {\rm Prune}(k|\*\alpha_0,0) =
 {\rm Prune}_{\rm WP}(k).
\end{align*}
This once more indicates that when the SP rule is satisfied, the WP rule must be satisfied as well.
When the RPB or RRPB sphere is used, the center of \ed{the} sphere is $\*q = \frac{\lambda_0 + \lambda_1}{2 \lambda_0} \*\alpha_0$.
%
Assuming that the solution for $\lambda_0$ is used as the reference solution, i.e., $\*\alpha = \*\alpha_0$, we obtain
\begin{align*}
 {\rm Prune}(k|\*q,r) 
 &=
 {\rm Prune}(k|\frac{\lambda_0 + \lambda_1}{2 \lambda_0} \*\alpha,r) \\
 &\geq
 {\rm Prune}(k|\frac{\lambda_0 + \lambda_1}{2 \lambda_0} \*\alpha,0) \\
 &= \frac{\lambda_0 + \lambda_1}{2 \lambda_0} {\rm Prune}(k| \*\alpha,0) \\
 &= \frac{\lambda_0 + \lambda_1}{2 \lambda_0} {\rm Prune}_{\rm WP}(k).
\end{align*}
Using this inequality, we obtain
\[
	{\rm Prune}(k|\*q,r) - {\rm Prune}_{\rm WP}(k) 
	\geq 
	\frac{\lambda_1 - \lambda_0}{2 \lambda_0} {\rm Prune}_{\rm WP}(k).
\]
From this inequality, if $\lambda_1 > \lambda_0$, then ${\rm Prune}(k|\*q,r) > {\rm Prune}_{\rm WP}(k)$ (note that ${\rm Prune}_{\rm WP}(k) \geq 0$ because $\*\alpha \geq \*0$), indicating that the pruning of WS is always tighter than that of the safe rule.
However, in our algorithm \ed{presented} in Section~\ref{sec:computations}, $\lambda_1 < \lambda_0$ holds because we start from a larger value of $\lambda$ and gradually decrease it.
Then, this inequality does not hold, and ${\rm Prune}(k|\*q,r) < {\rm Prune}_{\rm WP}(k)$ becomes a possibility.

When the WS and WP rules are strictly tighter than the SS and SP rules, respectively, using both of WS/WP and SS/SP rules is equivalent to using WS/WP only (except for dynamic screening).
%
Even in this case, the range-based safe approaches (the RSS and RSP rules) can still be effective.
When the range-based rules are evaluated, we obtain the range of $\lambda$ \ed{such that} the SS or SP rule is satisfied.
Thus, as long as $\lambda$ is in that range, we do not need to evaluate any safe or working set rules. 

%

\section{Algorithm and Computations}
\label{sec:computations}

\subsection{Training with Path-wise Optimization}

We employ \emph{path-wise optimization} \citep{friedman2007pathwise}, where the optimization starts from $\lambda = \lambda_{\max}$, \ed{which} gradually decreases $\lambda$ while optimizing $\*m$.
%
As \ed{can be seen from} \eqref{eq:lambdaMax}, $\lambda_{\max}$ is defined by the maximum of the inner product $\*C_{k,:}\*\alpha$. 
%
This value can also be found by \ed{a} tree search with pruning.
Suppose that we calculate $\*C_{k,:}\*\alpha$ while traversing the tree \ed{and} $\hat{\lambda}_{\max}$ is the current maximum value during the \ed{traversal}.
Using \ed{Lemma}~\ref{lmm:pruning}, we can derive the pruning rule
\[
\mathrm{Prune}(k|\bm\alpha,0)\le\hat{\lambda}_{\max}.
\]
%
If this condition holds, the descendant nodes of $k$ cannot be \ed{maximal}, and thus we can identify $\lambda_{\max}$ without calculating $\*C_{k,:}\*\alpha$ for all candidate features.

Algorithm~\ref{alg:RegularizationPath} shows the outer loop of our path-wise optimization. 
The {\tt TRAVERSE} and {\tt SOLVE} functions in Algorithm~\ref{alg:RegularizationPath} are shown in Algorithm~\ref{alg:Traverse} and \ref{alg:Solve}, respectively.
%
%
Algorithm~\ref{alg:RegularizationPath} first calculates $\lambda_{\max}$ which is the minimum $\lambda$ at which the optimal solution is $\*m^{\star} = \*0$ (line 3).
The outer loop in \ed{lines} 5-14 is the process of decreasing $\lambda$ with the decreasing rate $R$.
%
The {\tt TRAVERSE} function in line 7 determines the subset of features $\cF$ by traversing tree with \red{SS} and \red{WS}.
The inner loop (line 9-13) alternately solves the optimization problem with the current $\cF$ and updates $\cF$ until the duality gap becomes less than the given threshold ${\rm eps}$.

Algorithm~\ref{alg:Traverse} shows the {\tt TRAVERSE} function, which recursively visits \ed{tree nodes} to determine $\cF$.
The variable {\tt node.pruning} contains $\lambda'_a$ of RSP, and if the RSP condition \eqref{eq:RSP} is satisfied (line 3), the function returns the current $\cF$ (the node is pruned).
The variable {\tt node.screening} contains $\lambda_a$ of RSS, and if the RSS condition \eqref{eq:RSS} is satisfied (line 5), this node can be skipped, and the function proceeds to the next node.
%
If these two conditions are not satisfied, the function 
1) \ed{updates} {\tt node.pruning} and {\tt node.screening} if {\tt update} is true, 
and
2) \ed{evaluates} the conditions of RSP and WP (line 10), and RSS and WS (line 14).
%
%
\ed{In lines} 17-18, gSpan expands \ed{the} children of the current node, and for each child node, the {\tt TRAVERSE} function is called recursively.

Algorithm~\ref{alg:Solve} shows a solver for the primal problem with the subset of features $\cF$.
Although we employ a simple projected gradient algorithm, any optimization algorithm can be used in this process.
%
In \ed{lines} 7-10, the SS rule is evaluated at every \ed{after} ${\rm freq}$ iterations. 
Note that this SS is only for \ed{the} sub-problems \eqref{eq:primal-sub} and \eqref{eq:dual-sub} created by \ed{the} current $\cF$ (not for the original problems).

\begin{algorithm}[tbp]
	\caption{Path-wise Optimization}
	\label{alg:RegularizationPath}
	\Function($\textsc{PathwiseOptimization}{(}R, T, \mathrm{freq}, \mathrm{MaxIter}, \mathrm{eps}{)}$){
		$\bm m_0=\bm0, \bm\alpha_0=[2L,\ldots,2L,0,\ldots,0], \epsilon=0$\;
		$\lambda_{0}=\lambda_{\max}= \max_k \bm C_{k,:}\bm\alpha_0$ \Comment{compute $\lambda_{\max}$}
		Initialize root node as root.children = empty, root.screening = $\infty$, and root.pruning = $\infty$\;
		\For{\red{$t \in \{ 1, 2, \ldots, T \}$}}{
			$\lambda_t=R\lambda_{t-1}$\;
			$\mathcal{F}=\textsc{Traverse}(\lambda_{t-1}, \lambda_t, \bm\alpha_{t-1}, \epsilon, \mathrm{root}, \mathrm{true})$
			\Comment{get working set \& update range of $\lambda$}
			$\bm m_t=\bm m_{t-1}$\;
			\Repeat{$\frac{\mathrm{gap}}{P}\le \mathrm{eps}$}{
				$\bm m_t, \bm \alpha_t, P\! = \textsc{Solve}(\lambda_t, \bm m_t, \mathcal{F}, \mathrm{freq}, \mathrm{MaxIter}, \mathrm{eps})$\;
				$\mathcal{F}=\textsc{Traverse}(\mathrm{null}, \lambda_t, \bm\alpha_t, \mathrm{null}, \mathrm{root}, \mathrm{false})$
				\Comment{update working set}
				$\mathrm{gap}=P-D_{\lambda_t}^{\cF}(\bm\alpha_t)$\;
			}(\CommentHere{check optimality})
			$\epsilon=2\sqrt{\mathrm{gap}}$\;
		}
		\Return{$\{\bm m_t\}_{t=0}^{t=T}$}
	}
\end{algorithm}

\begin{algorithm}[tbp]
	\caption{Traverse gSpan with \red{RSSP}+\red{WSP}}
	\label{alg:Traverse}
	\Function($\textsc{Traverse}{(}\lambda_0, \lambda, \bm \alpha_0, \epsilon, \mathrm{node}, \mathrm{update}{)}$){
		$\mathcal{F}=\{ \}, k=\mathrm{node.feature}$\;
		\uIf(\CommentHere{RSP rule}){$\rm node.pruning\le \lambda$}{
			\textbf{return} $\mathcal{F}$\;
		}\uElseIf(\CommentHere{RSS rule}){$\rm node.screening \le \lambda$}{
			do nothing\;
		}\Else(\CommentHere{update the range of $\lambda$ if $\rm update=true$}){
			\If{$\rm update=true$}{
				$\mathrm{node.pruning}=\frac{\lambda_0(2\epsilon b+\|\bm\alpha_0\|_2b+a)}{2\lambda_0+\|\bm\alpha_0\|_2b-a}$ \Comment{eq.\eqref{eq:RSP}}
			}
			\If{$\rm node.pruning\le \lambda$ $\mathbf{or}$ $\mathrm{Prune}_{\rm WP}(k)\le \lambda$}{
				\Return{$\mathcal{F}$}
			}
			\If{$\rm update=true$}{
				$\mathrm{node.screening}=\frac{\lambda_0(2\epsilon\|\bm C_{k,:}\|_2+\|\bm\alpha_0\|_2\|\bm C_{k,:}\|_2+\bm C_{k,:}\bm\alpha_0)}{2\lambda_0+\|\bm\alpha_0\|_2\|\bm C_{k,:}\|_2-\bm C_{k,:}\bm\alpha_0}$ \Comment{eq.\eqref{eq:RSS}}
			}
			\If{$\rm node.screening > \lambda$ $\mathbf{and}$ $\bm C_{k,:}\bm\alpha_0> \lambda$}{
				$\mathcal{F}=\mathcal{F}\cup\{k\}$\;
			}
		}\red{
			\textsc{createChildren}(node)\;
			\For{\red{$\rm child \in node.children$}}{
				$\mathcal{F}=\mathcal{F}\cup\textsc{Traverse}(\lambda_0, \lambda, \bm \alpha_0, \epsilon, \mathrm{child}, \mathrm{update})$
			}}
		\Return{$\mathcal{F}$}
	}
	\Function($\textsc{createChildren}{(}\mathrm{node}{)}$){
		\If{$\rm node.children=empty$}{
			Set node.children by gSpan\;
			\For{$\rm child=node.children$}{
				child.children = empty\;
				child.screening = $\infty$, child.pruning = $\infty$
			}
		}
	}
\end{algorithm}
%
\begin{algorithm}[tbp]
	\caption{Gradient descent with dynamic screening}
	\label{alg:Solve}
	\Function($\textsc{Solve}{(}\lambda, \bm m, \mathcal{F}, \mathrm{freq}, \mathrm{MaxIter}, \mathrm{eps}{)}$
		\CommentHere{solve primal problem $P_\lambda^\mathcal{F}$, which is considered only for feature set $\mathcal{F}$}
		){
		\For{\red{$\rm iter \in \{ 0,1,\ldots,MaxIter \}$}}{
			Compute $\bm \alpha$ by \eqref{eq:primal2dual}\;
			$\mathrm{gap}=P_\lambda^\mathcal{F}(\bm m)-D_\lambda^\mathcal{F}(\bm\alpha)$\;
			\If(\CommentHere{convergence}){$\frac{\mathrm{gap}}{ P_\lambda^\mathcal{F}(\bm m) } \le \mathrm{eps}$}{
				\Return{$\bm m, \bm\alpha, P_\lambda^\mathcal{F}(\bm m)$}
			}
			\If{$\mathrm{mod}(\mathrm{iter}, \mathrm{freq}) = 0$}{
				\For(\CommentHere{perform dynamic screening}){$k\in\mathcal{F}$}{
					\If(\CommentHere{SS by DGB}){$\bm C_{k,:}\bm\alpha+2\sqrt{\mathrm{gap}}\|\bm C_{k,:}\|_2\le\lambda$}{
						$\mathcal{F}=\mathcal{F}-\{k\}$\;
					}
				}
			}
			$\bm m = [ \bm m - \gamma\nabla P_\lambda^\mathcal{F}(\bm m) ]_+$\Comment{update $\bm m$ ($\gamma$: step-size)}
		}
		\Return{$\bm m, \bm\alpha, P_\lambda^\mathcal{F}(\bm m)$}
	}
\end{algorithm}

\subsection{Enumerating Subgraphs for Test Data}

To obtain a feature vector for test data, we only need to enumerate subgraphs \ed{with} $m_k\ne 0$.
When gSpan is used as a mining algorithm, a unique code, called \emph{minimum DFS code}, is assigned to each node.
If a DFS code for a node is $(a_1, a_2, \ldots, a_n)$, a child node is represented by $(a_1, a_2, \ldots, a_n, a_{n+1})$.
%
This enables us to prune \ed{nodes that} does not generate subgraphs with $m_k\ne 0$.
%
Suppose that a subgraph $(a_1, a_2, a_3) = (x, y, z)$ \ed{must} be enumerated, and that we are \ed{currently at} node $(a_1) = (x)$.
%
Then, a child with $(a_1, a_2) = (x, y)$ should be traversed, but a child with $(a_1, a_2) = (x, w)$ cannot generate $(x,y,z)$, and consequently we can stop the \ed{traversal} of this node.

\subsection{Post-processing}
\label{ssec:post}

\subsubsection{Learning Mahalanobis Distance for Selected Subgraphs}
\label{sssec:mahalanobis}

Instead of $\*m$, the following Mahalanobis distance can be considered
\[ 
 d_{\*M}(\*x_i,\*x_j) 
 = (\*x_i - \*x_j)^\top \*M (\*x_i - \*x_j),
 \addtag \label{eq:mahalanobis}
\]
where $\*M$ is a positive definite matrix.
Directly optimizing $\*M$ requires $O(p^2)$ primal variables and semi-definite constraint, making the problem computationally expensive\ed{,} even for relatively small $p$.
%
\ed{Thus,} as optional post-processing, we consider optimizing the Mahalanobis distance \eq{eq:mahalanobis} for a small number of subgraphs selected by the optimized $\*m$.
Let $\cH \subseteq \cG$ be a set of subgraphs $m_{i(H)} > 0$ for $H \in \cH$ \ed{and} $\*z_i$ be a $h \coloneqq |\cH|$ dimensional feature vector consisting of $\phi_H(G_i)$ for $H \in \cH$.
For 
$\*M \in \RR^{h \times h}$,
we consider the following metric learning problem:
\[
 \begin{split}	 
  \min_{\*M \succeq \*O} &
  \sum_{i\in[n]}[
  \sum_{l\in \mathcal{D}_i}\ell_L(d_{\*M}(\*z_i,\*z_l))
  +\sum_{j\in \mathcal{S}_i}\ell_{-U}(-d_{\*M}(\*z_i,\*z_j))]
  + \lambda R(\*M).
 \end{split}
\]
\ed{Above,} $R: \RR^{h \times h} \rightarrow \RR$ is a regularization term for $\*M$, where a typical setting is $R(\*M) = {\rm tr} \*M + \frac{\eta}{2} \| \*M \|_F^2$ \ed{with} ${\rm tr}$ \ed{representing} the trace of a matrix.
This metric can be more discriminative, because it is optimized to the training data with a higher degree of freedom.

\subsubsection{Vector Representation of \ed{a} Graph}
\label{sssec:vector}

An explicit vector representation of an input graph can be obtained \ed{using} optimized $\bm m$ \ed{as follows}:
\[
 \bm x_i'=\sqrt{\bm m}\circ \bm x_i
 \addtag \label{eq:weighted-feature} 
\]
Unlike the original $\*x_i$, the new representation $\*x_i'$ is computationally tractable because of the sparsity of $\*m$, and simultaneously, 
this space should be highly discriminative.
This property is beneficial for further analysis of the graph data.
We show an example of applying the decision tree \ed{to} the learned space \ed{later in the paper.}

In the case of the general Mahalanobis distance \ed{given} in Section~\ref{sssec:mahalanobis}, we can obtain further transformation.
Let 
$\*M = \*V \*\Lambda \*V^\top$
be the eigenvalue decomposition of the learned $\*M$.
\ed{By} employing the regularization term $R(\*M) = {\rm tr} \*M + \frac{\eta}{2} \| \*M \|_F^2$, \ed{some} of the eigenvalues of $\*M$ can be shrunk to $0$ because ${\rm tr} \*M$ is equal to the sum of the eigenvalues.
If $\*M$ has $h' < h$ non-zero eigenvalues, $\*\Lambda$ can be written as a $h' \times h'$ diagonal matrix, and $\*V$ is a $h \times h'$ matrix \ed{such that} each column is the eigenvector of \ed{a} non-zero eigenvalue.
Then, a representation of \ed{the} graph is
\[
 \sqrt{\*\Lambda} \*V^\top \*z_i.
 \addtag \label{eq:transformation}
\]
This can be considered as a supervised dimensionality reduction from $h$- to $h'$-dimensional space.
Although each \ed{dimension no longer corresponds} to a subgraph in this representation, the interpretation remains clear because each dimension of the transformed vector is \ed{simply} a linear combination of $\*z_i$.

\section{Extensions}
\label{sec:extension}

In this section, we consider \ed{three} extensions of IGML: applications to other data types, employing a triplet loss function, and introducing vertex-label similarity.

\subsection{Application to Other Structured Data}
\label{sec:itemsetSequence}

\ed{In addition to graph data,} the proposed method can be applied to \ed{itemset}/sequence data . 
%
For \ed{an itemset}, the Jaccard index, defined as the size of the intersection of two sets divided by the size of the union, is the most popular similarity measure.
%
Although a few studies \ed{have considered} kernels for \ed{an itemset} \citep{zhang2007acik}, to \ed{the best of} our knowledge, it remains difficult to adapt a metric on \ed{a} given labeled dataset in an interpretable manner.
In contrast, there are many kernel approaches for sequence data. 
The spectrum kernel \citep{leslie2001spectrum} creates a kernel matrix by enumerating all $k$-length subsequences in the given sequence. 
The mismatch kernel \citep{leslie2004mismatch} enumerates subsequences allowing $m$ \ed{discrepancies} in a pattern of length $k$. 
The gappy kernel \citep{leslie2004fast,kuksa2008fast} counts the number of $k$-mers (subsequences) with a certain number of gaps $g$ that appear in the sequence. 
%
\ed{The above} kernels require the value of hyperparameter $k$, although various lengths may in fact \ed{be} related. 
%
%
The motif kernel \citep{zhang2006exmotif,pissis2013motex,pissis2014motex} counts the number of ``motifs'' appearing in the input sequences, \ed{the ``motif'' must be decided by the user.}
%
\ed{Because} these approaches are based on the idea of the `kernel', they are unsupervised, unlike our approach.

\begin{figure}[tbp]
	\includegraphics[width=\linewidth]{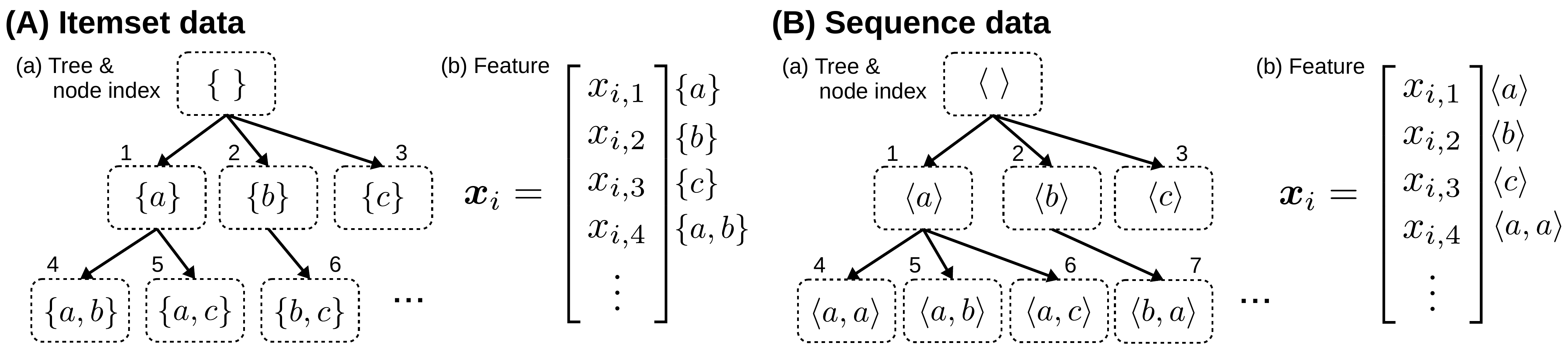}
	\caption{Schematic \ed{illustrations} of trees and features for \ed{(A) itemset and (B)} sequence data. }
	\label{fig:itemsetSequence}
\end{figure}

By employing a similar approach to the graph input, we can construct \ed{a} feature representation $\phi_H(X_i)$ for both \ed{itemset} and sequence data.
For the \ed{itemset} data, the $i$-th input is a set of items $X_i \subseteq \cI$, where $\cI$ is a set of all items\ed{, e.g.,}
%
$X_1 = \{ a, b \}, X_2 = \{ b, c, e \}, \ldots$ with the candidate items $\cI = \{a, b, c, d, e\}$.
%
The feature $\phi_H(X_i)$ is defined by $1_{H \subseteq X_i}$ for $\forall H \subseteq \cI$.
%
This feature $\phi_H(X_i)$ also has \ed{monotonicity} $\phi_{H^\prime}(X_i) \leq \phi_{H}(X_i)$ for $H^\prime \supseteq H$.
In \ed{sequence data}, the $i$-th input $X_i$ is a sequence of items.
Thus, the feature $\phi_H(X_i)$ is defined from the frequency of a sub-sequence $H$ in the given $X_i$.
%
For example, if we have $X_i = \langle b, b, a, b, a, c, d \rangle$ and $H = \langle b, a \rangle$, then $H$ occurs \ed{twice} in $X_i$.
For \ed{sequence data}, the monotonicity property is \ed{again} guaranteed \ed{because}
$\phi_{H^\prime}(X_i) \leq \phi_{H}(X_i)$\ed{,}
where $H$ is a sub-sequence of $H^\prime$.
Because of these monotonicity properties, we can apply the same pruning procedures to both of \ed{itemset} and sequence data.
%
\figurename~\ref{fig:itemsetSequence} shows \ed{examples} of \ed{trees that} can be constructed by \ed{itemset and} sequence mining algorithms \citep{agrawal1994fast,pei2001prefixspan}.

\subsection{Triplet Loss}
\label{sssec:triplet}

We \ed{formulate} the loss function of IGML as the pair-wise loss \eqref{eq:loss}. 
%
\ed{Triplet loss functions are also} widely used in metric learning \citep[e.g.,][]{weinberger2009distance}: 
\[
\sum_{(i,j,l)\in\mathcal{T} }\ell_1(\bm m^\top \bm c_{il}-\bm m ^\top\bm c_{ij}), 
\]
where $\cal T$ is an index set of triplets consisting of $(i,j,l)$ \ed{satisfying} $y_i=y_j, y_i\ne y_l$. 
This loss incurs a penalty when the distance between samples in the same class is larger than the distance between samples in different classes. 
Because the loss is defined by a `triplet' of samples, this approach can be more time-consuming than the \ed{pairwise} approach.
%
In contrast, the relative evaluation such as $d_{\bm m}(\bm x_i,\bm x_j) < d_{\bm m}(\bm x_i,\bm x_l)$ (the $j$-th sample must be closer to the $i$-th sample \ed{than} the $l$-th sample) can capture the \ed{higher-order relations between} input objects rather than penalizing the pair-wise distance.

A pruning rule can be derived even for the case of \ed{triplet loss}. 
%
By defining $\bm c_{ijl}\coloneqq\bm c_{il}-\bm c_{ij}$, the loss function \ed{can be} written as 
\[
\sum_{(i,j,l)\in\mathcal{T} }\ell_1(\bm m^\top \bm c_{ijl}). 
\]
%
Because this \ed{has} the same form as \ed{pairwise} loss with $L=1$ \ed{and} $U=0$, the optimization problem is reduced to the same form as the pairwise case. 
%
We \ed{require} a slight modification \ed{of Lemma}~\ref{lmm:pruning} because of the change of the constant coefficients (i.e., from $\bm c_{ij}$ to $\bm c_{ijl}$).
The equation \eqref{eq:pruning} is changed to
\[
\mathrm{Prune}(k | \bm q, r)\coloneqq
\sum_{(i,j,l)\in\cal T}q_{ijl}\max\{x_{i,k},x_{l,k}\}^2+r\sqrt{\sum_{ijl}\max\{x_{i,k},x_{l,k}\}^4 }. 
 \addtag \label{eq:triplet-pruning}
\]
This is easily proven using
\[
		c_{ijl, k'}=(x_{i,k'}-x_{l,k'})^2-(x_{i,k'}-x_{j,k'})^2\le \max\{x_{i,k}, x_{l,k}\}^2, \forall k' \supseteq k,
\] 
which is an immediate consequence of the monotonicity inequality \eqref{eq:monotonicity}.
%


\subsection{Considering Vertex-Label Similarity}
\label{sssec:similarity}

\begin{figure}[t]
 \centering
 \includegraphics[width=0.35\linewidth]{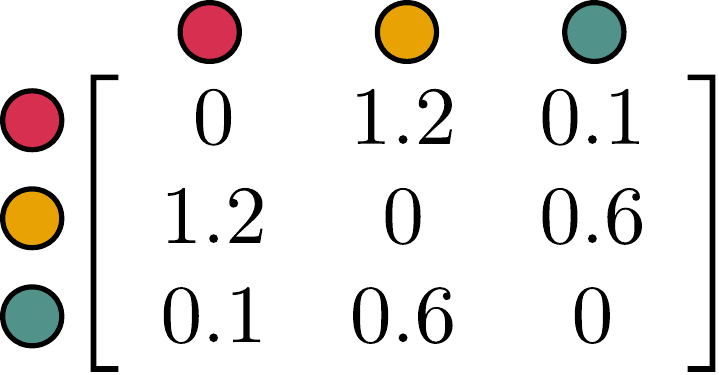}
 \caption{
 \ed{Dissimilarity} matrix among vertex-labels. 
 }
 \label{fig:sim-mat}
\end{figure}

\ed{Because} IGML is based on the exact matching of subgraphs to create the feature $\phi_H(G)$, it is difficult to provide a prediction \ed{for} a graph that does not exactly match many of the selected subgraphs.
Typically, this happens when the test dataset has a different distribution of vertex-labels.
%
For example, in the case of the prediction on a chemical compound group whose atomic compositions are largely different from those \ed{in} the training dataset, the exact match may not be expected as in the case of the training dataset.
\ed{Therefore,} we consider incorporating similarity/dissimilarity information of graph vertex-labels \ed{to} relax this exact matching constraint.
A toy example of vertex-label dissimilarity is shown in \figurename~\ref{fig:sim-mat}.
In this case, the `red' vertex is similar to the `green' vertex, while it is dissimilar to the `yellow' vertex.
%
For example, we can create this type of table by using prior domain knowledge (e.g., chemical properties of atoms).
%
%
Even when no prior information is available, a similarity matrix can be inferred \ed{using} any embedding \ed{method} \citep[e.g.,][]{huang2017label}.
%
%

\begin{figure}[t]
 \centering
 \includegraphics[width=\linewidth]{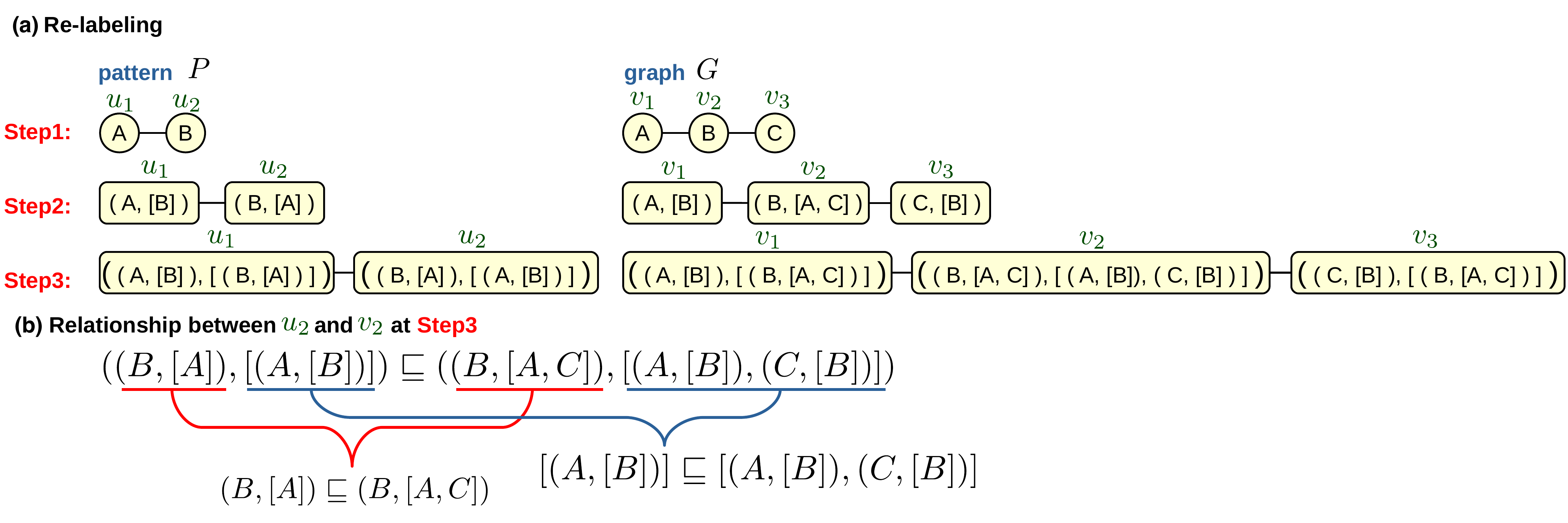}
 \caption{
 Re-labeling and inclusion relationship. 
 $(X, [\,])$ is abbreviated as $X$, where $X\in\{A,B,C\}$. 
 (a) In each step, all vertices are re-labeled by combining a vertex-label and neighboring labels at the previous step.
 (b) Example of inclusion relationship, defined by \eq{eq:label-relation} and \eq{eq:label-relation-def}.
 The relation $L_P(u_2,3) \sqsubseteq L_G(v_2,3)$ is satisfied between $u_2$ and $v_2$ at Step3.
 }
 \label{fig:wl-extend}
\end{figure}

Because it is difficult to directly incorporate similarity information into our \ed{subgraph isomorphism-based} feature $\phi_H(G)$, we first introduce a relaxed evaluation of inclusion of a graph $P$ in a given graph $G$.
We assume that $P$ is obtained from the gSpan tree of \ed{the} training data.
%
%
Our approach is based on the idea of \ed{`re-labeling' graph} vertex-labels in the Weisfeiler-Lehman (WL) kernel \citep{shervashidze2011weisfeiler}, which is a well-known graph kernel with \ed{an approximate graph} isomorphism test.
%
%
%
\figurename~\ref{fig:wl-extend} (a) shows an example of the re-labeling procedure, which is performed \ed{in a} fixed number of recursive steps. 
The number of steps is denoted as $T$ ($T = 3$ in the figure) \ed{and} is assumed to be pre-specified.
%
%
In step $h$, each graph vertex $v$ has a \emph{level $h$ hierarchical label}
$L_G(v,h) \coloneqq (F^{(h)}, S^{(h)} = [S^{(h)}_1, \ldots, S^{(h)}_n])$, 
where 
$F^{(h)}$ 
is recursively defined by the level $h - 1$ hierarchical label of the same vertex, 
i.e., 
$F^{(h)} = L_G(v,h-1)$, 
and 
$S^{(h)}$ 
is a multiset created by the level $h - 1$ hierarchical labels $L_G(v',h-1)$ from \ed{all} neighboring vertices $v'$ connected to $v$.
Note that a multiset, denoted by `$[,]$', is a set where duplicate elements are allowed.
For example, in the graph $G$ shown \ed{on} the right side of \figurename~\ref{fig:wl-extend} (a), the hierarchical label of the vertex $v_1$ on \ed{level} $h = 3$ is $L_G(v_1,3) = ((A, [B]), [(B, [A,C])])$.
In this case, 
$F^{(3)} = (A, [B])$, which is equal to $L_G(v_1,2)$,
and
$S^{(3)}_1 = (B, [A,C])$, which is equal to $L_G(v_2,2)$.
The original label $A$ can also be regarded as a hierarchical label $(A,[\,])$ on the level $h = 1$, but it is shown as `$A$' for simplicity.

%

We define a relation of the inclusion `$\sqsubseteq$' between two 
hierarchical labels 
$L_{P}(u,h) = (F^{(h)}, S^{(h)} = [S^{(h)}_1, \ldots, S^{(h)}_m])$
and 
$L_{G}(v,h) = (F^{\prime (h)}, S^{\prime (h)} = [S^{\prime (h)}_1, \ldots, S^{\prime (h)}_n])$,
which originate from the two vertices $u$ and $v$ in graphs $P$ and $G$, respectively.
%
%
We say that
$L_{P}(v,h)$
is included in
$L_{G}(u,h)$ \ed{and denote it} by
\begin{equation}
	L_{P}(v,h) \sqsubseteq L_{G}(u,h)
	\label{eq:label-relation}
\end{equation}
when the following recursive condition is satisfied\ed{:}\\
\begin{subequations}
 \label{eq:label-relation-def}
\begin{empheq}[left=\empheqlbrace]{alignat=3}
 & F^{(h)} = F^{\prime (h)},	&& \text{ if } S^{(h)} = S^{\prime (h)} = [\,], \label{eq:label-relation-a}
 \\
 & F^{(h)} \sqsubseteq F^{\prime (h)} \land \exists \sigma (\land_{i \in [m]} S^{(h)}_i \sqsubseteq S_{\sigma(i)}^{\prime (h)}),
 && \text{ otherwise}, 
 \label{eq:label-relation-b}
\end{empheq}
\end{subequations}
%
where $\sigma: [m] \rightarrow [n]$ is an injection from $[m]$ to $[n]$ (i.e., $\sigma(i) \ne \sigma(j)$ when $i \ne j$), and
$\exists \sigma (\land_{i \in [m]} S^{(h)}_i\sqsubseteq S^{\prime (h)}_{\sigma(i)})$
indicates that there exists an injection $\sigma$ \ed{that} satisfies 
$S^{(h)}_i\sqsubseteq S^{\prime (h)}_{\sigma(i)}$ 
for $\forall i \in [m]$. 
The first condition \eqref{eq:label-relation-a} is for the case of 
$S^{(h)} = S^{\prime (h)} =[\,]$, 
which occurs at the first level $h=1$, and in this case, it simply evaluates whether the two hierarchical labels are equal\ed{, i.e.,} $F^{(h)} = F^{\prime (h)}$.
Note that when $h = 1$, the hierarchical label is simply $(X,[])$, where $X$ is one of the original vertex-labels.
In the other case \eqref{eq:label-relation-b}, 
both of the two conditions 
$F^{(h)} \sqsubseteq F^{\prime (h)}$ 
and 
$\exists \sigma (\land_{i \in [m]} S^{(h)}_i\sqsubseteq S^{\prime (h)}_{\sigma(i)})$
are recursively defined.
Suppose that we already evaluated the level $h - 1$ relation 
$L_{P}(u,h-1) \sqsubseteq L_{G}(v,h-1)$ 
for \ed{all} pairs $\forall (u,v)$ from $P$ and $G$.
%
Because $F^{(h)} = L_{P}(u,h-1)$ and $F^{\prime (h)} = L_{G}(v,h-1)$, the condition
$F^{(h)} \sqsubseteq F^{\prime (h)}$
is equivalent to 
$L_{P}(u,h-1) \sqsubseteq L_{G}(v,h-1)$, 
which is assumed to be already obtained on the level $h-1$ computation. 
%
Because $S^{(h)}_i$ and $S^{\prime (h)}_i$ are also from hierarchical labels on \ed{level} $h - 1$,
the condition
$\exists \sigma (\land_{i \in [m]} S^{(h)}_i \sqsubseteq S^{\prime (h)}_{\sigma(i)})$
is also recursive.
%
From the result of the level $h - 1$ evaluations, we can \ed{determine} whether $S^{(h)}_i \sqsubseteq S^{\prime (h)}_{j}$ holds for $\forall (i,j)$. 
Then, the evaluation of the condition 
$\exists \sigma (\land_{i \in [n]} S^{(h)}_i \sqsubseteq S^{\prime (h)}_{\sigma(i)})$
is reduced to a matching problem from $i \in [m]$ to $j \in [n]$. 
%
%
This problem can be simply transformed into a \emph{maximum bipartite matching} problem for a pair of  
$\{ S^{(h)}_1, \ldots, S^{(h)}_n \}$ 
and
$\{ S^{\prime (h)}_1,\ldots,S^{\prime (h)}_m \}$,
where edges exist on a set of pairs 
$\{ (i,j) \mid S^{(h)}_i\sqsubseteq S^{\prime (h)}_j \}$.
%
When \ed{the maximum number of matchings} is equal to $m$, this means that there exists an injection $\sigma(i)$ \ed{satisfying} $\land_{i \in [m]} S^{(h)}_i\sqsubseteq S^{\prime (h)}_{\sigma(i)}$.
%
It is well known that the maximum bipartite matching can be reduced to \ed{the \emph{maximum flow problem}}, \ed{which} can be solved \ed{in} the polynomial time \citep{goldberg1988new}. 
An example of the inclusion relationship is shown in \figurename~\ref{fig:wl-extend} (b).
Let $|P|$ and $|G|$ be the numbers of vertices in $P$ and $G$\ed{, respectively}.
Then, multisets of the level $T$ hierarchical labels of all the vertices in $P$ and $G$ are written as
$[L_{P}(u_i,T)]_{i \in [|P|]} \coloneqq [L_{P}(u_1,T), L_{P}(u_2,T), \ldots, L_{P}(u_{|P|},T)]$
and 
$[L_{G}(v_i,T)]_{i \in [|G|]} \coloneqq [L_{G}(v_1,T), L_{G}(v_2,T), \ldots, L_{G}(v_{|G|},T)]$, 
respectively.
For a feature of a given input graph $G$, we define the \emph{approximate subgraph isomorphism feature (ASIF)} as follows\ed{:}
\begin{equation}
 \label{eq:ASIF}
  x_{P \sqsubseteq G} 
  \coloneqq 
  \begin{cases}
   1, & \text{ if }
   \exists 
   \sigma
   (\land_{i \in [ |P| ]} L_{P}(u_i,T) \sqsubseteq L_{G}(v_{\sigma(i)},T) ),
   \\
   0, & \text{ otherwise. }
  \end{cases} 
\end{equation}
This feature approximately evaluates the existence of a subgraph $P$ in $G$ using the level $T$ hierarchical labels.
ASIF satisfies the monotone decreasing property \eqref{eq:monotonicity}, i.e., 
$x_{P'\sqsubseteq G}\le x_{P\sqsubseteq G}$ if $P' \sqsupseteq P$, 
because the number of conditions in \eqref{eq:label-relation-def} only increases when $P$ grows. 
%


To incorporate \ed{label} dissimilarity information (\ed{as shown in} \figurename~\ref{fig:sim-mat}) into ASIF, 
%
we first extend the label inclusion relation \eq{eq:label-relation} by using \ed{the} concept of \emph{optimal transportation cost}. 
%
As a \ed{label similarity-based} relaxed evaluation of 
$L_{P}(v,h) \sqsubseteq L_{G}(u,h)$, 
we define an asymmetric cost between
$L_P(u,h)$
and
$L_G(v,h)$
as follows
\begin{subequations}
 \label{eq:label-cost}
 \begin{empheq}[ left = {%
 \begin{aligned}[b]
  & \mathrlap{ \mathrm{cost}_h(L_P(u,h)\! \rightarrow\! L_G(v,h))} \\[5ex]
  & \coloneqq 
 \end{aligned}
  \empheqlbrace}]
  {alignat=2}
	& \mathrm{dissimilarity} (F^{(h)}, F^{\prime (h)}),  & \text{ if } S^{(h)} \!=\! S^{\prime (h)} \!=\! [\,], \label{eq:label-cost-a}
	\\
	& \mathrm{cost}_{h-1} (F^{(h)} \!\to\! F^{\prime (h)}) +  \nonumber \\ 
	& \ \mathrm{LTC}(S^{(h)} \!\to\! S^{\prime (h)}, \mathrm{cost}_{h-1}),	& \text{ otherwise},   
   \label{eq:label-cost-b}
  \end{empheq}
\end{subequations}
where the second term of \eq{eq:label-cost-b} is
\begin{equation}
	\mathrm{LTC}(S^{(h)} \to S^{\prime (h)}, \mathrm{cost}_{h-1}) 
	\coloneqq 
	\min_{ \sigma \in \cI } \sum_{i \in [m]} \mathrm{cost}_{h-1}(S^{(h)}_i \to S^{\prime (h)}_{\sigma(i)}),
	\label{eq:otc}
\end{equation}
which we refer to as the \emph{label transportation cost} (LTC) representing the optimal transportation from the multiset $S^{(h)}$ to another multiset $S^{\prime (h)}$ among the set of all injections 
$\cI \coloneqq \{ \forall \sigma: [m] \rightarrow [n] \mid \sigma(i) \neq \sigma(j) \text{ for } i \neq j \}$.
%
The equation \eq{eq:label-cost} has a recursive structure \ed{similar to that of} \eq{eq:label-relation}.
%
The first case \eqref{eq:label-cost-a} occurs when $S^{(h)} = S^{\prime (h)} =[\,]$, which \ed{is} at the first level $h = 1$.
In this case, $\mathrm{cost}_1$ is defined by $\mathrm{dissimilarity}(F^{(1)},F^{\prime (1)})$, which is directly obtained as a dissimilarity between original labels since $F^{(1)}$ and $F^{\prime (1)}$ stem from the original vertex-labels.
In the other case \eqref{eq:label-cost-b}, the cost is recursively defined as the sum of \ed{the} cost from $F^{(h)}$ to $F^{\prime (h)}$ and the optimal-transport cost from $S^{(h)}$ to $S^{\prime (h)}$.
%
Although \ed{this} definition is recursive, as in the case of ASIF, the evaluation can be performed by computing sequentially from $h = 1$ to $h = T$.
Because 
$F^{(h)} = L_{P}(v,h-1)$ 
and
$F^{\prime (h)} = L_{G}(u,h-1)$, 
the first term $\mathrm{cost}_{h-1}(F^{(h)} \to F^{\prime (h)})$ represents the cost between hierarchical labels on the level $h-1$, which is assumed to \ed{already have been} obtained.
%
%
The second term $\mathrm{LTC}(S^{(h)} \to S^{\prime (h)}, \mathrm{cost}_{h-1})$ evaluates the best match between $[S^{(h)}_1, \ldots, S^{(h)}_m]$ and $[S^{\prime (h)}_1, \ldots, S^{\prime (h)}_n]$, as defined in \eq{eq:otc}.
This matching problem can be seen as an optimal transportation problem, which minimizes the cost of the transportation of $m$ items to $n$ warehouses under the given cost matrix specified by $\mathrm{cost}_{h-1}$.
%
The values of $\mathrm{cost}_{h-1}$ for all the pairs in $[m]$ and $[n]$ are also available from the computation \ed{at} the level $h - 1$.
%
For the given cost values, the problem \ed{of} $\mathrm{LTC}(S^{(h)} \to S^{\prime (h)}, \mathrm{cost}_{h-1})$ can be reduced to \ed{a} \emph{minimum-cost-flow problem} on a bipartite graph with a weight $\mathrm{cost}_{h-1}(S^{(h)}_i \to S^{\prime (h)}_j, \mathrm{cost}_{h-1})$ between $S^{(h)}_i$ and $S^{\prime (h)}_j$, \ed{which} can be solved in polynomial time \citep{goldberg1988new}.
We define an asymmetric transport cost for two graphs $P$ and $G$, which we call the \emph{graph transportation cost} (GTC), as LTC from \ed{all} level $T$ hierarchical labels of $P$ 
to those of $G$:
\[
	\mathrm{GTC}(P \to G)
	\coloneqq
	\mathrm{LTC}(
	[L_{P}(u_i,T)]_{i \in [|P|]}
	\rightarrow
	[L_{G}(v_i,T)]_{i \in [|G|]},
	\mathrm{cost}_{T}
	).
\]
Then, as a feature of the input graph $G$, we define the following \emph{sim-ASIF}:
\begin{equation}
	\label{eq:sim-ASIF}
	x_{P \to G} \coloneqq \exp\{-\rho \,\mathrm{GTC}(P \to G)\},  
\end{equation}
where $\rho > 0$ is a hyperparameter.
This sim-ASIF can be regarded as a generalization of \eqref{eq:ASIF} based on the vertex-label similarity.
When $\mathrm{dissimilarity}(F^{(1)}, F^{\prime (1)}) \coloneqq \infty \times 1_{F^{(1)} \ne F^{\prime (1)}}$, the feature \eqref{eq:sim-ASIF} is equivalent to \eqref{eq:ASIF}. 
Similarly to ASIF, $\mathrm{GTC}(P\to G)$ satisfies the monotonicity property
\[
	\mathrm{GTC}(P \to G) \le \mathrm{GTC}(P' \to G)~\text{for}~P' \sqsupseteq P
\]
because the number of vertices to transport increases as $P$ grows.
Therefore, sim-ASIF \eqref{eq:sim-ASIF} satisfies the monotonicity property\ed{, i.e.,} $x_{P'\to G}\le x_{P\to G}$ if $P'\sqsupseteq P$. 


%
From the definition \eq{eq:sim-ASIF}, sim-ASIF always has a positive value $x_{P \rightarrow G} > 0$ except \ed{when} $\mathrm{GTC}(P \to G) = \infty$, which may not be suitable for identifying a small number of important subgraphs.  
Further, in sim-ASIF, the bipartite graph in the minimum-cost-flow calculation
$\mathrm{LTC}(S \to S', \mathrm{cost}_{h-1})$
is always a complete bipartite graph, where \ed{all} vertices in $S$ are connected to \ed{all} vertices in $S'$.
Because the efficiency of most of standard minimum-cost-flow algorithms depends on the number of edges, this may \ed{entail} a large computational cost.
%
As an extension \ed{to mitigate} these issues, a threshold can be introduced into sim-ASIF as follows:
\begin{equation}
	\label{eq:threshold-feature}
	x\coloneqq
	\begin{cases}
	\exp\{-\rho \,\mathrm{GTC}(P \to G)\}, &\exp\{-\rho \,\mathrm{GTC}(P \to G)\}>t\\
	0, &\exp\{-\rho \,\mathrm{GTC}(P \to G)\}\le t
	\end{cases}, 
\end{equation}
where $t > 0$ is a threshold parameter.
In this definition, 
\ed{$x = 0$} 
when 
$\exp\{-\rho \,\mathrm{GTC}(P \to G)\}\le t$,
i.e., 
$\mathrm{GTC}(P \to G) \ge -(\log t)/\rho$. 
This indicates that if a cost is larger than $-(\log t)/\rho$, we can regard the cost as $\infty$. 
Therefore, at any $h$, if the cost between $S^{(h)}_i$ and $S^{\prime (h)}_j$ is larger than $-(\log t)/\rho$, the edge between $S^{(h)}_i$ and $S^{\prime (h)}_j$ is not necessary.
Then, the number of matching pairs can be less than $m$ in $\mathrm{LTC}(\cdot)$ because of the lack of edges, and in this case, the cost is regarded as $\infty$. 
%
Furthermore, if $\mathrm{cost}_h (F^{(h)} \to F^{\prime (h)})$ is larger than $-(\log t)/\rho$ in \eqref{eq:label-cost-b}, the computation of $\mathrm{LTC} (S^{(h)} \to S^{\prime (h)}, \mathrm{cost}_{h-1})$ is not \ed{required} because $x=0$ is determined. 
Note that \ed{transportation-based graph metrics have} been studied \citep[e.g.,][]{vay2019optimal}, but the purpose of \ed{such studies was} to evaluate the similarity between two graphs (not inclusion).
Our (sim-)ASIF provides a feature with the monotonicity property as a natural relaxation of subgraph isomorphism, by which the optimality of our pruning strategy can be guaranteed. 
In contrast, there \ed{have been} many studies \ed{on} inexact graph matching \citep{yan2016short} such as eigenvector\ed{-} \citep{leordeanu2012unsupervised,kang2013fast}, edit distance\ed{-} \citep{gao2010survey}, and random walk\ed{-based} \citep{gori2005exact,cho2010reweighted} methods.
%
Some of \ed{these methods} provide a score \ed{for} the matching\ed{,} which can be seen as a similarity score between a searched graph pattern and a matched graph.
%
However, \ed{they do} not guarantee \ed{the} monotonicity of the similarity score for pattern growth. 
If the similarity score satisfies monotonicity, it can be combined with IGML.
%
Although we only \ed{consider vertex-labels, edge-labels} can also be incorporated into (sim-)ASIF.
A simple approach is to transform a labeled-edge into a labeled-node with two unlabeled edges, such that (sim-)ASIF is directly applicable.

\section{Experiments}
\label{sec:experiments}

We evaluate the performance of IGML using the benchmark datasets shown in \tablename~\ref{tbl:f-score}. 
These datasets \ed{are available} from \citet{KKMMN2016}. 
We did not use \ed{edge labels} because \ed{the} implementations of compared methods cannot deal with \ed{them}, and the maximum connected graph is used if the graph is not connected.
\red{
Note that IGML currently cannot directly deal with continuous attributes, so we did not use them. 
%
%
A possible approach would be to perform discretization or quantization before the optimization, such as  taking grid points or applying clustering in the attribute space.
%
Building a more elaborated approach, such as dynamically determining discretization, is a possible future directions.
}
%
\ed{The} \#maxvertices \ed{column} in the table \ed{indicates} the size (number of vertices) of the maximum subgraph considered in IGML.
\red{
To fully identify important subgraphs, a large value of \#maxvertices is preferred, but this can cause a correspondingly large memory requirement to store the gSpan tree. 
%
For each dataset, we set the largest value for which IGML could finish with a tractable amount memory.
}
The sets $\mathcal{S}_i$ and $\mathcal{D}_i$ \ed{were} selected as the ten nearest neighborhoods of $\bm x_i$ ($K=|\mathcal{S}_i|=|\mathcal{D}_i|=10$) by using the WL-Kernel.
%
A sequence of the regularization coefficients \ed{was} created by equally \ed{spacing} 100 grid points on \ed{a} logarithmic scale between $\lambda_{\max}$ and $0.01\lambda_{\max}$.
%
\red{
We set the minimum support in gSpan as $0$, meaning that all the subgraphs in a given dataset were enumerated for as far as the graph satisfies the \#maxvertices constraint.
%
The gSpan tree is mainly traversed when the beginning of each $\lambda$ as shown in Algorithm~\ref{alg:RegularizationPath} (in the case of WS-based approaches, the tree is also traversed at every working set update).
Note that the tree is dynamically constructed during this traversal without constructing the entire tree beforehand.
}
%
%
In the working-set method, after convergence, it is necessary to traverse the tree again in order to confirm the overall optimality. 
%
If the termination condition is not satisfied, optimization with a new working set must be performed. 
%
The termination condition for the optimization is that the relative duality gap is less than $10^{-6}$.
%
In \ed{the} experiment, we used $g(x) = 1_{x > 0}$ in $\phi_H(G)$ with Lemma~\ref{lmm:pruning-binary} unless otherwise noted.
The dataset \ed{was} randomly divided in such a way that the ratio of partitioning \ed{was} $\rm train:validation:test = 0.6:0.2:0.2$, and our experimental \ed{results were averaged over 10 runs}.

\subsection{Evaluating Computational Efficiency}
\label{ssec:efficiency}

In this section, we confirm the effect of \ed{the} proposed pruning methods. 
We evaluated four settings\ed{:}
\red{Safe Screening and Pruning: ``\red{SSP}'', Range based Safe Screening and Pruning: ``\red{RSSP}'', Working set Selection and Pruning: ``\red{WSP}'', and the combination of WSP and RSSP: ``\red{WSP}+\red{RSSP}''. }
%
Each method \ed{performed} dynamic screening with DGB at every update of $\bm m$. 
We here used the AIDS dataset, where \#maxvertices=30.
In this dataset, when we fully \ed{traversed} the gSpan tree without safe screening/working set selection, the number of tree nodes was more than $9.126 \times 10^7$\ed{, at which point} our implementation with gSpan stopped because we ran out of memory.

\begin{figure}[tbp]
	\subfloat[]{
 \includegraphics[width=.46\linewidth]{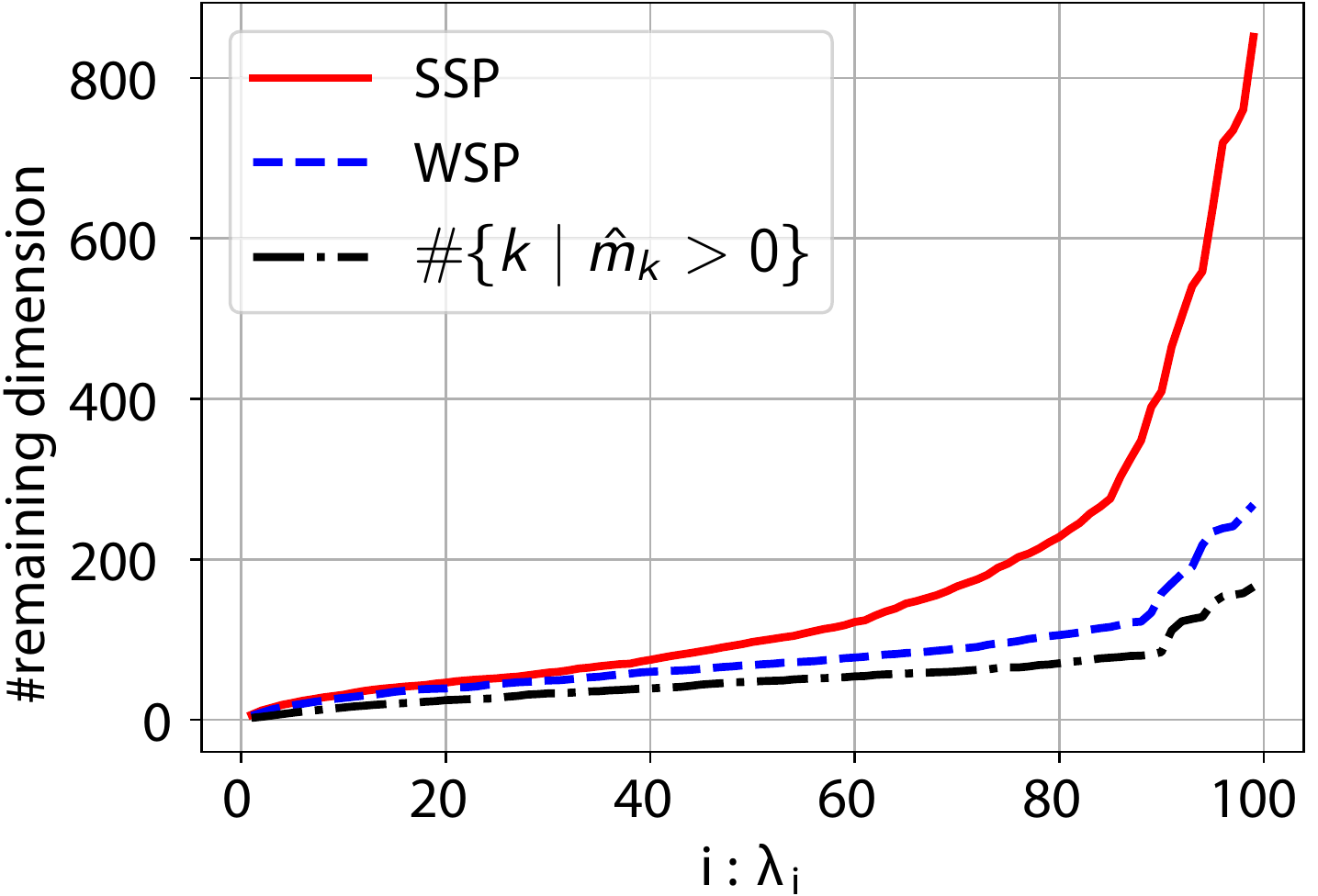}
		\label{fig:aids_remained}
	}
	\subfloat[]{
\includegraphics[width=.485\linewidth]{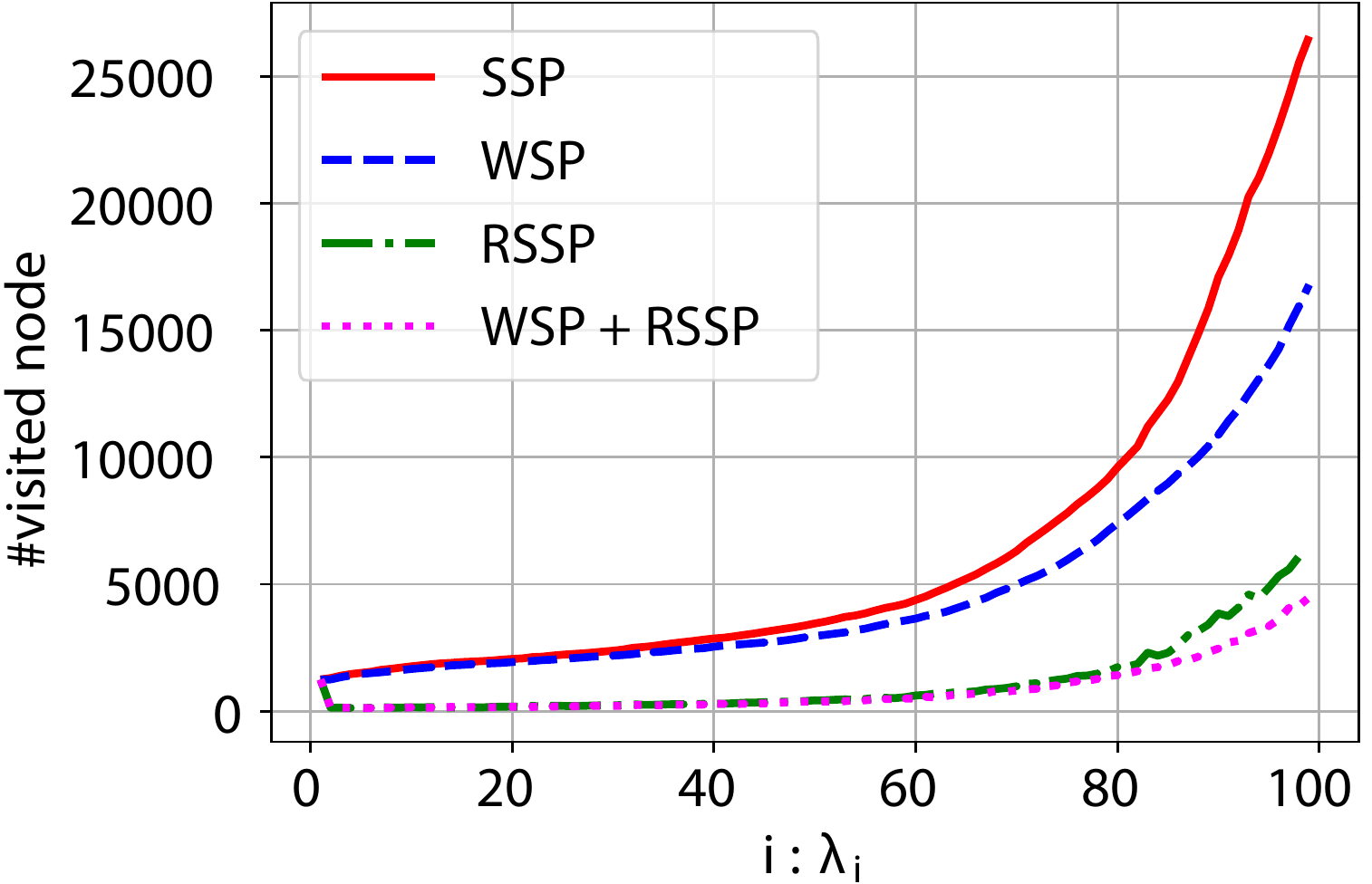}
		\label{fig:aids_visit}
	}
	\caption{
		\ed{(a)} Size of $\cF$, and \ed{(b)} number of visited nodes.
		%
		Both \ed{were} evaluated at the first \ed{traversal} of each $\lambda$, where the index is shown \ed{on} the horizontal axis.
		%
		The dataset employed here \ed{was} AIDS.
	}
	\label{fig:aids}
\end{figure}

\figurename~\ref{fig:aids} \subref{fig:aids_remained} shows the size of $\cF$ after the first traverse at each $\lambda$, and the number of non-zero $m_k$ after the optimization is also shown as a baseline.
%
%
We first observe that both approaches \ed{significantly} reduced the number of features.
%
Even for the largest case, where \ed{approximately} 200 of features were finally selected by $m_k$, only less than \ed{1000} features remained.
%
We observe that \red{WSP} \ed{exhibited} significantly smaller values than \red{SSP}.
%
Instead, \red{WSP} may need to perform the entire tree search again because it cannot guarantee the sufficiency of \ed{the} current $\cF$\ed{,} while \red{SSP} does not need to search the tree again because it guarantees that $\cF$ must contain all $m_k \neq 0$.

%
The number of visited nodes in the first \ed{traversal} at each $\lambda$ is shown in \figurename~\ref{fig:aids}~\subref{fig:aids_visit}.
\red{
Here, we added RSSP and WSP+RSSP, which are not shown in \figurename~\ref{fig:aids} \subref{fig:aids_remained}.
Note that the \#remaining dimensions is same for SSP and RSSP, and for WSP and WSP+RSSP.
Because RSSP is derived from SSP, it does not change the number of screened features.
As we discussed in Section~\ref{sssec:relation-with-safe-rules}, WSP removes more features than RSSP, though it is not safe. 
}
%
We \ed{observed} that the \#visited nodes of \red{SSP} \ed{was} the largest, but it \ed{was} less than \ed{approximately} 27000 ($27000 / 9.126 \times 10^7 \approx 0.0003$).
%
Comparing \red{SSP} and \red{WSP}, 
%
we see that \red{WSP} pruned a larger number of nodes.
%
In contrast, the \#visited nodes of \red{RSSP} \ed{was} less than 6000. 
The difference between \red{SSP} and \red{RSSP} indicates that a larger number of nodes can be skipped by the range\ed{-}based method. 
%
Therefore, by combining the node skip by \red{RSSP} with \ed{the} stronger pruning of \red{WSP}, the \#visited nodes was further reduced.
%
\red{
RSSP and WSP+RSSP had larger values at $\lambda_0$ than the subsequent $\lambda_i$. 
This is because of the effect of range-based screening and pruning.
At $\lambda_0$, every visited node in the tree calculates the ranges in which the screening and pruning rules are satisfied (i.e., RSS and RSP rules), and as a result, some nodes can be skipped during that $\lambda_i$ is in those ranges.
%
At every $\lambda_i$ for $i > 0$, the ranges are updated only in the (non-skipped) visited nodes, and thus, the range-based rules take the effect except for $\lambda_0$.
}


The total time \ed{of} the path-wise optimization is shown in \tablename~\ref{tbl:time}. 
%
\red{RSSP and WSP+RSSP} were fast with regard to the traversing time, and \red{WSP and WSP+RSSP} were fast with regard to the solving time. 
\red{
Note that because the tree is dynamically constructed during the traverse, the `Traverse' time includes the time spent on the tree construction.
}
In total, \red{WSP}+\red{RSSP} was \ed{the} fastest. 
%
\ed{These results indicate} that our method only took \ed{approximately} $1$ minute to solve the optimization problem with more than $9.126 \times 10^7$ variables.
We also show the computational cost evaluation for other datasets in the Appendix~\ref{app:cpu-time}.





\red{
Although we have confirmed that IGML works efficiently on several benchmark datasets, completely elucidating general complexity of IGML is remains as future work.
The practical complexity at least depends on the graph size in the training data, \#maxvertices, \#samples, and the pruning rate.
In terms of the graph size, traversing a large graph dataset using gSpan can be intractable because it requires all matched subgraphs to be maintained at each tree node.
Therefore, applying IGML to large graphs, e.g., graphs with more than thousands of nodes, would be difficult.
%
Meanwhile, the scalability of IGML depends not only on the sizes of graphs but also strongly on the performance of the pruning.
%
However, we still do not have any general analytic complexity evaluation for the rate of the pruning that avoids exponential worst-case computations.
%
%
%
In fact, we observed that there exist datasets in which efficiency of the pruning is not sufficient.
For example, on the IMDB-BINARY and IMDB-MULTI datasets, which are also from \citep{KKMMN2016}, a large number of small subgraphs are shared across all the different classes and instances (i.e., $x_{i,k} = 1$ for $\forall i$).
Our upper bound in the pruning is based on the fact that 
$x_{i,k'} \leq x_{i,k}$
for descendant node $k'$ in the mining tree.
This bound becomes tighter when $x_{i,k} = 0$ for many $i$ because $0$ is the lower bound of $x_{i,k}$.
%
In contrast, when many instances have $x_{i,k} = 1$, the bound can be loose, making traversal intractable.
This is an important open problems common in predictive mining methods \citep{nakagawa2016safe,morvan2018whinter}.
}

\begin{table}[tbp]
	\centering
	\caption{Total time in path-wise optimization (sec) on AIDS dataset. }
	\label{tbl:time}
	{\footnotesize
	\hdashlinewidth=0.2mm
	\hdashlinegap=0.5mm
		\begin{tabular}{c|r:r:r}
		Method~\textbackslash~Process	
				&Traverse	&Solve		&Total		\\
				\hline 
		\red{SSP}	&25.9		&			&112.7 \\
				&$\pm$4.0	&86.7		&$\pm$16.5	\\
				\cdashline{1-2} \cdashline{4-4}
		\red{RSSP}&7.7		&$\pm$14.1	&94.4	\\
				&$\pm$1.6	&			&$\pm$15.1	\\
				\hline
		\red{WSP}	&39.1		&			&94.1		\\
				&$\pm$3.7	&\B{55.0}	&$\pm$11.6	\\
				\cdashline{1-2}\cdashline{4-4}
		\red{WSP}+	&\B{7.4}	&$\pm$12.1	&\B{62.5}	\\
		\red{RSSP}&$\pm$1.1	&			&$\pm$12.3
		\end{tabular}
	}
\end{table}
\subsection{Predictive Accuracy Comparison}
\label{ssec:accuracy-comparison}

In this section, we compare the prediction accuracy of IGML with \ed{those of} the Graphlet-Kernel (GK)\citep{shervashidze2009efficient}, Shortest-Path Kernel (SPK)\citep{borgwardt2005shortest}, Random-Walk Kernel (RW)\citep{vishwanathan2010graph}, Weisfeiler-Lehman Kernel (WL)\citep{shervashidze2011weisfeiler}, and Deep Graph Convolutional Neural Network (DGCNN)\citep{zhang2018end}.
We \ed{used} the \ed{implementations} available at \ed{the} URLs \ed{in the footnote}\footnote{
\red{\url{https://github.com/ysig/GraKeL} for GK, 
\url{http://mlcb.is.tuebingen.mpg.de/Mitarbeiter/Nino/Graphkernels/} for the other graph kernel, 
and \url{https://github.com/muhanzhang/pytorch_DGCNN} for DGCNN.}}.
%
\red{
Note that we mainly compared methods for obtaining a metric between graphs. 
%
%
%
%
The graph kernel approach is one of most important existing approaches to defining a metric space of non-vector structured data.
%
%
Although kernel functions are constructed in an un-supervised manner, their high prediction performance has been widely shown.
%
In particular, the WL kernel is known for its comparable classification performance to recent graph neural networks \citep[e.g.,][]{niepert2016learning,morris2019weisfeiler}.
%
Meanwhile, DGCNN can provide a vector representation of an input graph by using the outputs of some middle layer, which can be interpreted that a metric space is obtained through a supervised learning.
%
%
We did not compare with \citep{saigo2009gboost, nakagawa2016safe, morvan2018whinter} as they only focused on specific linear prediction models rather than building a general discriminative space.
}
%
%
\red{
We employed the $k$-nearest neighbors ($k$-nn) classifier to directly evaluate the discriminative ability of feature spaces constructed by IGML and each kernel function. 
%
We here employed the $k$-nn classifier for directly evaluating discriminative ability of feature spaces constructed by IGML and each kernel function.
}
\red{
A graph kernel can be seen as an inner-product $k(G_i,G_j) = \langle \varphi(G_j), \varphi(G_j) \rangle$, where $\varphi$ is a projection from a graph to reproducing kernel Hilbert space.
Then, the distance can be written as
$\| \varphi(G_j) - \varphi(G_j)  \| = \sqrt{k(G_i,G_i) - 2 K(G_i,G_j) + k(G_j,G_j)}$.
}
The values of $k$ \ed{for} the $k$-nn \ed{were} $k=1, 3, 5, 7, ..., 49$ and hyperparameters of each method \ed{were} selected \ed{using} the validation data, and the prediction accuracy \ed{was} evaluated on the test data. 
The graphlet size \ed{for} GK \ed{was set} \red{up to 6}.
The parameter $\lambda_{\rm RW}$ \ed{for} RW \ed{was set to} the recommended $\lambda_{\rm RW}=\max_{i\in\mathbb{Z}: 10^i<1/d^2}10^i$, where $d$ denotes the maximum degree.
%
The loop parameter $h$ of WL \ed{was} selected from $0,1,2,...,10$ by using the validation data.
%
\ed{For} DGCNN, the number of hidden units and their sort-pooling \ed{were} also selected \ed{using} the validation data, each ranging from $64, 128, 256$ and from $40\%, 60\%, 80\%$, respectively.

The micro-F1 score for each dataset is shown in \tablename~\ref{tbl:f-score}. 
``IGML (Diag)'' \ed{indicates} IGML with the weighted squared distance \eqref{eq:sq-distance}, and ``IGML (Diag$\to$Full)'' indicates that \red{with} post-processing \ed{using} the Mahalanobis distance \eqref{eq:mahalanobis}. 
%
\red{``IGML (Diag)'' yielded the best or comparable to the best score on seven out of nine datasets.
This result is impressive because IGML uses a much simpler metric than the other methods.
%
Among the seven datasets, ``IGML (Diag$\to$Full)'' slightly improved the mean accuracy on four datasets, but the difference was not significant.
This may suggest that the diagonal weighting can have enough performance in many practical settings.
WL kernel also exhibited superior performance, showing the best or comparable to the best accuracy on six datasets.}
%
DGCNN \ed{showed} high accuracy with \ed{on the DBLP\_v1 dataset}, which has a large number of samples, while \ed{its accuracy was low for the other datasets}.

\begin{table*}[tbp]
 \centering
 \caption{
 Comparison of micro-F1 score. OOM means out of memory. ``$>$1week'' indicates that the algorithm \ed{ran} for more than a week.
 %
 \ed{The numbers after the ``$\pm$'' are} the standard \ed{deviations}. 
 %
 Every dataset \ed{had} two classes. 
 \red{The bold font indicates the best average value and the `*' symbol indicates that it is comparable to the best value in terms of one-sided t-test (significance level $0.05$).}
 }
 \label{tbl:f-score}
 {
	\scriptsize
	\hdashlinewidth=0.2mm
	\hdashlinegap=0.5mm
	\tabcolsep = 3pt
	\begin{tabular}{c|rrrrrrrrrr}
		\multirow{2}{*}{Method\,\textbackslash\,Dataset}	&\multirow{2}{*}{AIDS}		&\multirow{2}{*}{BZR}		&\multirow{2}{*}{DD}			&\multirow{2}{*}{DHFR}		&FRANKE	
										&Mutag		&\multirow{2}{*}{NCI1}		&\multirow{2}{*}{COX2}		&\multirow{2}{*}{DBLP\_v1}	\\
										&			&			&			&			&NSTEIN
										&enicity	&			&			&			
		\\
		\hline
		\#samples						&2000		&405		&1178		&467		&4337			
										&4337			&4110		&467		&19456		
		\\
		\#maxvertices					&30			&15			&30			&15			&15				
										&10				&15			&15			&30			
		\\
		\hline
		GK								
& \red{     0.967} & \red{     0.794} & \red{     0.700} & \red{     0.707} & \red{     0.628} & \red{     0.667} & \red{     0.640} & \red{     *0.781} &         OOM 
\\
& \red{$\pm$     0.010} & \red{$\pm$     0.036} & \red{$\pm$     0.015} & \red{$\pm$     0.034} & \red{$\pm$     0.012} & \red{$\pm$     0.010} & \red{$\pm$     0.016} & \red{$\pm$     0.026} &     
\\
		\hdashline
		SPK								&0.994 & \red{*}0.842	&$>$1week	&0.737	&0.640
										&0.719		&0.722	& \red{*}0.774	&0.784
		\\
										&$\pm$0.003&$\pm$0.039	&	&$\pm$0.040	&$\pm$0.012
										&$\pm$0.014		&$\pm$0.012	&$\pm$0.034	&$\pm$0.012
		\\ \hdashline
		RW								&\B{0.998}&0.811&OOM		&0.659	&0.616
										&0.679		&0.649	& \red{*}0.770	&OOM		
		\\
										&$\pm$0.002&$\pm$0.025&		&$\pm$0.032	&$\pm$0.013
										&$\pm$0.018		&$\pm$0.017	&$\pm$0.038	&		
		\\ \hdashline
		WL								&0.995	& \red{*}0.854 & \red{*}0.769	& \red{*}0.780	& \red{*}0.694
										&0.768	& \red{*}0.772	&\B{0.790}&0.814
		\\
										&$\pm$0.003	&$\pm$0.039	&$\pm$0.027	&$\pm$0.045	&$\pm$0.017
										&$\pm$0.012	&$\pm$0.015	&$\pm$0.040&$\pm$0.014
		\\ \hdashline
		DGCNN							&0.985	&0.791	& \red{*}0.773	&0.678	&0.615
										&0.705	&0.706& \red{*}0.764	&\B{0.927}
		\\
									&$\pm$0.005	&$\pm$0.020	&$\pm$0.023	&$\pm$0.030	&$\pm$0.016
										&$\pm$0.018	&$\pm$0.016&$\pm$0.039	&$\pm$0.003
		\\
		\hline
		IGML (Diag)					&0.976	&\B{0.860}& \red{*}0.778&\B{0.797}	& \red{*}0.696
										& \red{*}0.783 & \red{*}0.775	& \red{*}0.777	&0.860
		\\
							&$\pm$0.006	&$\pm$0.030&$\pm$0.026&$\pm$0.035	&$\pm$0.014
										&$\pm$0.016&$\pm$0.012	&$\pm$0.037	&$\pm$0.005
		\\ \hdashline
		IGML 		    &0.977	&0.830	&\B{0.783}& \red{*}0.794&\B{0.699}
										&\B{0.790}	&\B{0.782}& \red{*}0.773	&0.856
		\\
		(Diag$\to$Full)	&$\pm$0.008	&$\pm$0.029	&$\pm$0.022&$\pm$0.042&$\pm$0.013
										&$\pm$0.023	&$\pm$0.014&$\pm$0.038	&$\pm$0.005
		\\
	\end{tabular}
	}
\end{table*}

\subsection{Illustrative Examples of Selected Subgraphs}

\begin{figure}[tbp]
 \includegraphics[width=\linewidth]{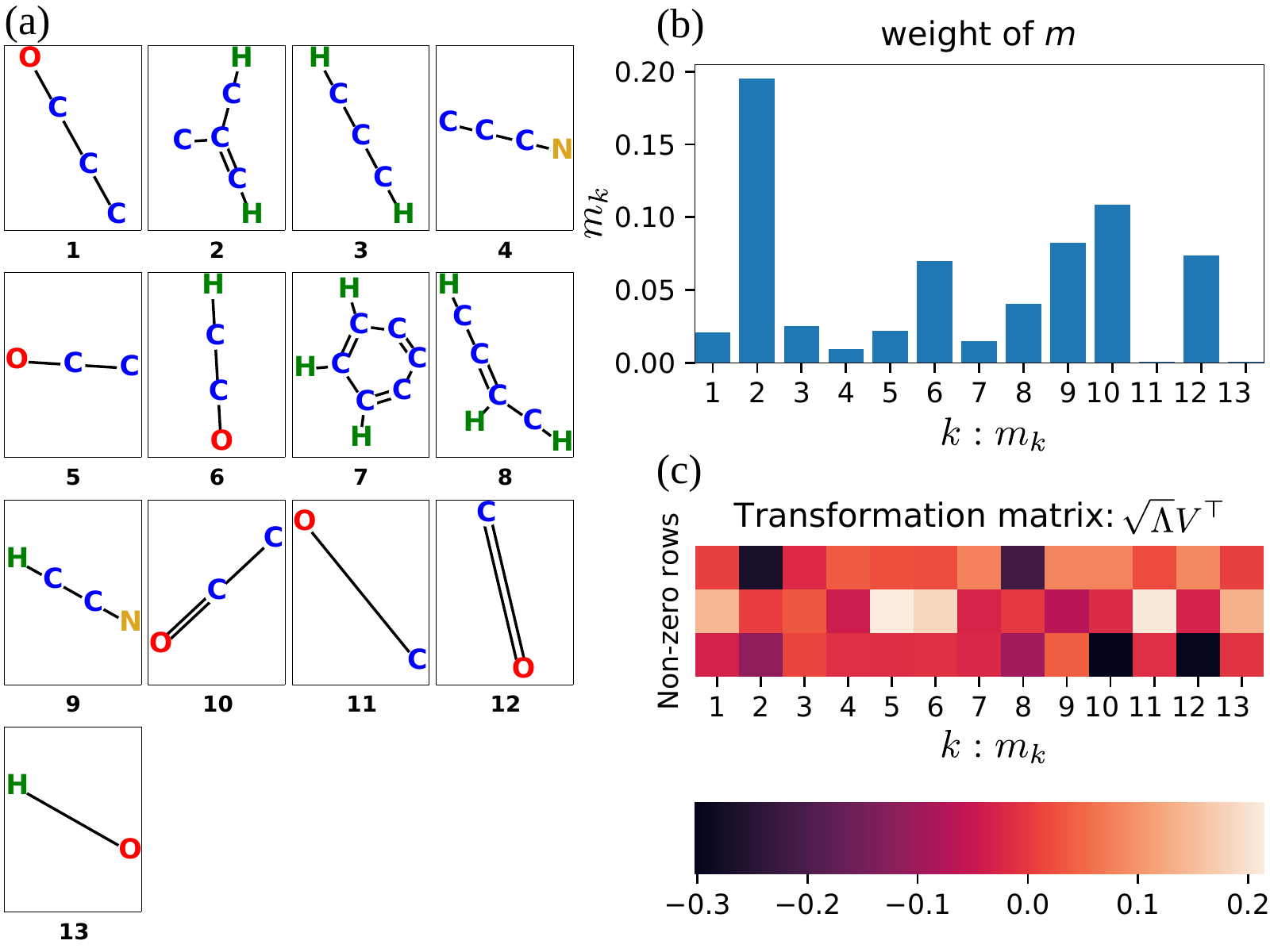}
 \caption{
 \ed{Examples} of selected subgraphs. 
 (a): \ed{Illustrations} of subgraphs. 
 (b): Learned \ed{weights} of \ed{subgraphs}. 
 (c): \ed{Transformation} matrix \ed{heatmap} \eqref{eq:transformation}.
 }
	\label{fig:lam6summary}
\end{figure}

\figurename~\ref{fig:lam6summary} shows an illustrative example of IGML on the Mutagenicity dataset, {where mutagenicity \ed{was} predicted from a graph representation of molecules.}
%
%
\figurename~\ref{fig:lam6summary}~(a) is a graphical representation of subgraphs, each of which has a weight shown in (b).
For example, we can clearly see that \ed{subgraph} \#2 is estimated as an important sub-structure to discriminate different classes.
%
\figurename~\ref{fig:lam6summary}~(c) shows a heatmap of the transformation matrix $\sqrt{\bm\Lambda}\bm V^\top$ optimized for \ed{the} thirteen features, containing three non-zero eigenvalues.
%
For example, we see that \ed{the subgraphs of} \#10 and \#12 have similar columns in the heatmap.
%
This indicates that these two similar subgraphs (\#10 contains \#12) are shrunk to almost same representation \ed{by} the regularization term $R(\*M)$.

\begin{figure}[tbp]
	\includegraphics[width=\linewidth]{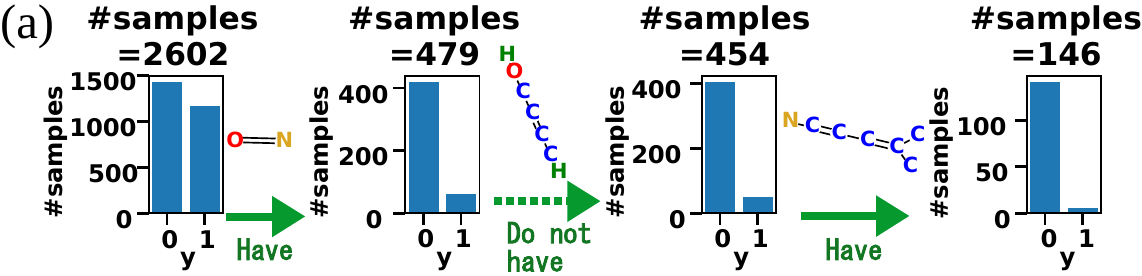}\\
	\includegraphics[width=\linewidth]{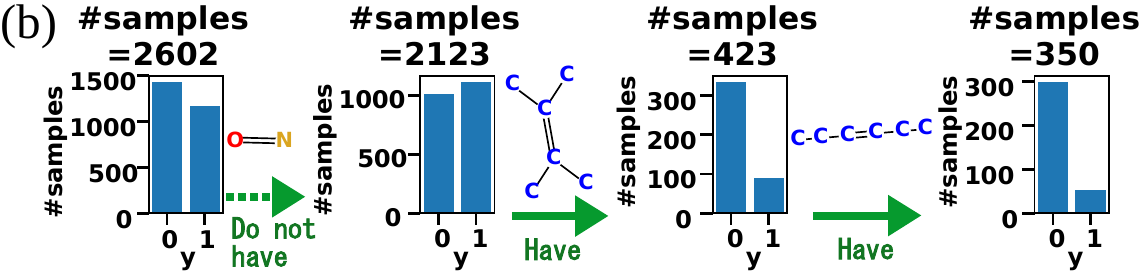}
	\caption{
	Examples of paths on decision tree constructed by selected subgraphs. 
	\#samples indicates \ed{the} number of samples satisfying all preceding conditions.
	}
	\label{fig:lam99barTreeGridPath}
\end{figure}

As another example of graph data analysis on the learned representation, we applied the decision tree algorithm to the obtained feature \eqref{eq:weighted-feature} on the Mutagenicity dataset.
%
Although there \ed{has been} a study constructing \ed{a} decision tree directly for graph data \citep{nguyen2006constructing}, it requires a severe restriction on the \ed{patterns} to be considered for computational \ed{feasibility}. 
%
In contrast, \ed{because} \eqref{eq:weighted-feature} is a simple vector representation with a reasonable dimension, it is quite easy to apply the decision tree algorithm.
%
\ed{We selected} two paths from the obtained decision tree as shown in \figurename~\ref{fig:lam99barTreeGridPath}.
For example, in the path (a), if a given graph contains ``$\red{\bm O}\!=\!\yellow{\bm N}$'', and does not contain ``$\green{\bm H}\!-\!\red{\bm O}\!-\!\blue{\bm C}\!-\!\blue{\bm C}\!=\!\blue{\bm C}\!-\!\blue{\bm C}\!-\!\green{\bm H}$'', and contains 
\renewcommand{\arraystretch}{0.85}
``$\yellow{\bm N}\!-\!\blue{\bm C}\!=\!\blue{\bm C}\!-\!\blue{\bm C}\!=\!\blue{\bm C}\!\!<\!\!\begin{array}{l}\blue{\bm C}\\\blue{\bm C}\end{array}$'', 
\renewcommand{\arraystretch}{1}
the given graph is predicted as $y=0$ with probability $140/146$. 
Both rules clearly separate the two classes, which is highly insightful as we can trace the process of the decision based on the subgraphs.

\subsection{Experiments for Three Extensions}
\label{ssec:exp-ext}

In this section, we \ed{evaluate} the performance of the three extensions of IGML described in Section~\ref{sec:extension}.

\begin{figure}[tbp]
 \includegraphics[width=\linewidth]{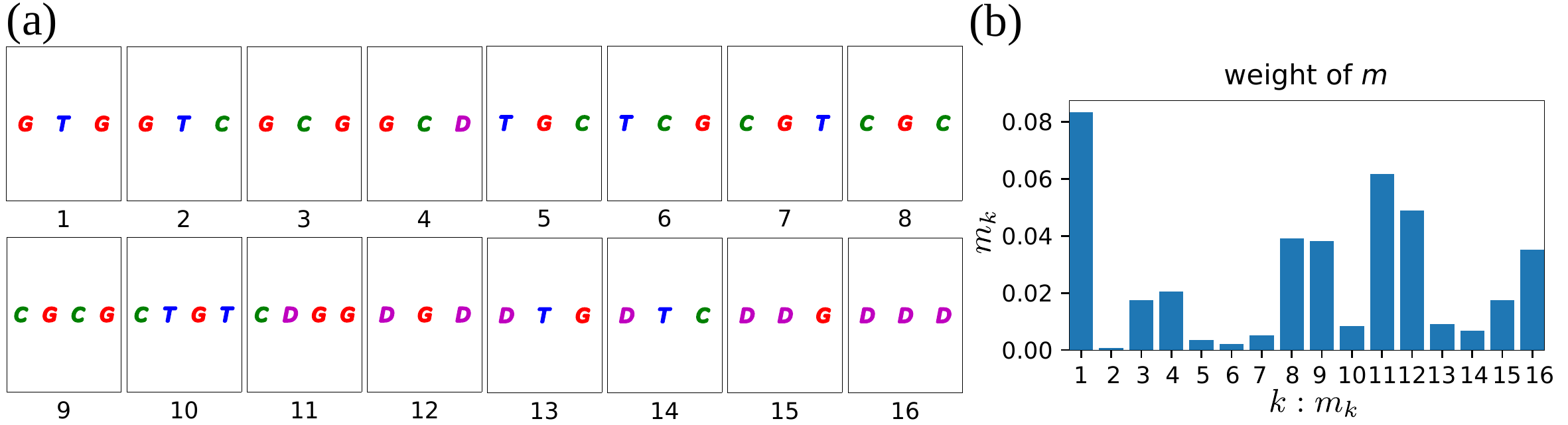}
 \caption{
 Examples of (a) selected sequences and (b) their weights \ed{for the} promoters dataset.
 }
 \label{fig:promoters}
\end{figure}

First, we \ed{evaluated} the performance of IGML \ed{on itemset and} sequence data \ed{using} the benchmark datasets shown in the first two rows \ed{of} \ed{\tablename~\ref{tbl:itemset} and \ref{tbl:sequence}}. 
These datasets can be obtained from \citep{Dua:2019} and \citep{CC01a}, \ed{respectively}.
%
We set the maximum-pattern size considered \ed{by} IGML as 30. 
%
\tablename~\ref{tbl:itemset} \ed{lists} the micro-F1 \ed{scores} on the \ed{itemset datasets}. 
%
We used $k$-nn with the Jaccard similarity as a baseline, \ed{where} $k$ was selected \ed{using} the validation set\ed{, as described} in Section~\ref{ssec:accuracy-comparison}.
The scores of both of IGML (Diag) and (Diag$\rightarrow$Full) were superior to those of the Jaccard similarity on \ed{all} datasets. 
%
\tablename~\ref{tbl:sequence} \ed{lists} the micro-F1 \ed{scores} on the sequence dataset. 
Although IGML (Diag) did not outperform the mismatch kernel \citep{leslie2004mismatch} for the promoters dataset, IGML (Diag$\rightarrow$Full) achieved a higher F1-score than the kernel on \ed{all} datasets.
\figurename~\ref{fig:promoters} shows an illustrative example of sequences \ed{identified} by IGML on the promoters dataset, where the task \ed{was} to predict whether an input DNA sequence \ed{stems} from a promoter region.
%
\figurename~\ref{fig:promoters}~(a) is a graphical representation of the sequence, \ed{and} the corresponding weights \ed{are} shown in (b). 
For example, the sub-sequence \#1 in (a) can be considered as an important sub-sequence to discriminate different classes.

\begin{table}[tbp]
	\centering
	\caption{Micro-F1 \ed{scores} on \ed{itemset} \ed{datasets}. }
	\label{tbl:itemset}
	\footnotesize
	\begin{tabular}{c|rrr}
		Method~\textbackslash~Dataset	&dna			&car			&nursery		\\
		\hline
		\#samples						&2000			&1728			&12960			\\
		\hline
		Jaccard Similarity				&0.860$\pm$0.017&0.888$\pm$0.020&0.961$\pm$0.006\\
		\hline
		IGML (Diag)					&0.908$\pm$0.014&0.936$\pm$0.011&0.982$\pm$0.005\\
		IGML (Diag$\to$Full)		&\B{0.931}$\pm$0.009&\B{0.948}$\pm$0.014&\B{0.993}$\pm$0.002\\
	\end{tabular}
\end{table}

\begin{table}[tbp]
	\centering
	\caption{Micro-F1 \ed{scores} on sequence \ed{datasets}. }
	\label{tbl:sequence}
	\footnotesize
	\begin{tabular}{c|rr}
		Method~\textbackslash~Dataset	&promoters		&splice			\\
		\hline
		\#samples						&106			&3190			\\
		\hline
		Mismatch Kernel					&0.832$\pm$0.081&0.596$\pm$0.017\\
		\hline
		IGML (Diag)					&0.800$\pm$0.104&0.651$\pm$0.015\\
		IGML (Diag$\to$Full)		&\B{0.886}$\pm$0.068&\B{0.694}$\pm$0.017\\
	\end{tabular}
\end{table}

Second, we show \ed{the} results of the triplet formulation described in Section~\ref{sssec:triplet}. 
To create the triplet set $\mathcal{T}$, we followed the approach \ed{in} \citet{shen2014efficient}, where $k$ neighborhoods in the same class $\bm x_j$ and $k$ neighborhoods in different classes $\bm x_l$ \ed{were} sampled for each $\bm x_i$ ($k=4$). 
Here, IGML with the \ed{pairwise} loss is referred to as `IGML (Pairwise)', and IGML with the \ed{triplet} loss is referred to as `IGML (Triplet)'.
\tablename~\ref{tbl:ext-score} compares the micro-F1 \ed{scores} of IGML (Pairwise) and IGML (Triplet).
\ed{IGML} (Triplet) showed higher F1-scores than IGML (Pairwise) \ed{on} three \ed{of} nine datasets, but it was not computable on the two datasets due to \ed{running} out of memory (OOM). 
This is because the pruning rule \ed{in} the triplet case \eqref{eq:triplet-pruning} was looser than \ed{in} the pair-wise case. 
%
\begin{table}
 \centering
 \caption{
 Comparison of \ed{IGML (Pairwise) and IGML (Triplet) in terms of} micro-F1 score.
 }
	\label{tbl:ext-score}
	{
	\scriptsize
	\hdashlinewidth=0.2mm
	\hdashlinegap=0.5mm
	\tabcolsep = 3pt
	\begin{tabular}{c|rrrrrrrrrr}
		\multirow{2}{*}{Method\,\textbackslash\,Dataset}	&\multirow{2}{*}{AIDS}		&\multirow{2}{*}{BZR}		&\multirow{2}{*}{DD}			&\multirow{2}{*}{DHFR}		&FRANKE	
										&Mutag		&\multirow{2}{*}{NCI1}		&\multirow{2}{*}{COX2}		&\multirow{2}{*}{DBLP\_v1}	\\
										&			&			&			&			&NSTEIN
										&enicity	&			&			&			
										\\
		\hline
		IGML (Pairwise)						&\B{0.976}	&\B{0.860}	&\B{0.778}	&0.797	&\B{0.696}	
											&0.783		&0.775		&\B{0.777}	&\B{0.860}
		\\
		from \tablename~\ref{tbl:f-score}	&$\pm$0.006	&$\pm$0.030&$\pm$0.026&$\pm$0.035	&$\pm$0.014
											&$\pm$0.016&$\pm$0.012	&$\pm$0.037	&$\pm$0.005
		\\  \hdashline
		IGML (Triplet)						&0.968		&0.844		&OOM		&\B{0.811}	&0.693
											&\B{0.808}	&\B{0.782}	&0.765		&OOM	
		\\
		\#maxvertices=10					&$\pm$0.012	&$\pm$0.032	&			&$\pm$0.033	&$\pm$0.013
											&$\pm$0.012	&$\pm$0.013	&$\pm$0.042	&			
	\end{tabular}
	}
\end{table}

Finally, we \ed{evaluated the} sim-ASIF \eqref{eq:threshold-feature}. 
We set the scaling factor of the exponential function as $\rho=1$, the threshold of the feature as $t=0.7$, and the number of re-labeling steps as $T=3$. 
%
\ed{We} employed a simple heuristic approach to create a dissimilarity matrix among vertex-labels \ed{using} labeled graphs in the given dataset.
%
Suppose that \ed{the} set of possible vertex-labels is $\cL$, and $f(\ell,\ell')$ is the frequency that $\ell \in \cL$ and $\ell' \in \cL$ are adjacent in all graphs of the dataset.
%
\red{By} concatenating $f(\ell,\ell')$ for all $\ell' \in \cL$, we \ed{obtained} a vector representation of a label $\ell$.
We \ed{normalized} this vector representation such that the vector \ed{had} the unit L2 norm.
By calculating the Euclidean distance of this normalized representations, we \ed{obtained} the dissimilarity matrix of vertex-labels.
%
We are particularly interested in the case where the distribution of \ed{the} vertex-label frequency is largely different between the training and test datasets, \ed{because} in this case the exact matching of IGML may not be suitable to provide \ed{a} prediction.
%
We synthetically \ed{emulated} this setting by splitting \ed{the} training and test datasets \ed{using} a clustering algorithm.
%
Each input graph \ed{was} transformed into a vector created by the frequencies of \ed{each vertex-label} $\ell \in \cL$ contained in that graph.
\red{Subsequently, we applied the $k$-means clustering to split the dataset into two clusters, for which ${\cal C}_1$ and ${\cal C}_2$ denote sets of assigned data points, respectively.
%
We used ${\cal C}_1$ for the training and validation datasets and ${\cal C}_2$ is used as the test dataset, where $|{\cal C}_1| \geq |{\cal C}_2|$. 
Following the same partitioning policy as in the above experiments, the size of the validation data was set as the same size of $\cC_2$, resulting from which the size of the training set was $|{\cal C}_1| - |{\cal C}_2|$.
}
%
\tablename~\ref{tbl:k-means-datasplit} \ed{lists} the comparison of the micro-F1 scores on the AIDS, Mutagenicity, and NCI1 datasets.
\red{
We did not consider other datasets as their training set sizes created from the above procedure were too small.
}
%
%
We fixed the \#maxvertices of sim-ASIF \ed{to} 8, which \ed{was} less than the value in our original IGML evaluation \tablename~\ref{tbl:f-score}, because sim-ASIF takes more time than the feature without vertex-label similarity.
For the original IGML, we show the result \ed{for the setting} in \tablename~\ref{tbl:f-score} and the results with \#maxvertices 8.
%
IGML with sim-ASIF was superior to the original IGML for the both \#maxvertices settings on the AIDS and NCI1 datasets, although it has smaller \#maxvertices settings, as shown in \tablename~\ref{tbl:k-means-datasplit}.
%
\ed{On} the Mutagenicity dataset, sim-ASIF was inferior to the original IGML \ed{reported in} \tablename~\ref{tbl:f-score}, but in the comparison under the same \#maxvertices value, their \ed{performances were} comparable.
%
These results suggest that when the exact matching of the subgraph is not appropriate, sim-ASIF can improve the prediction performance of IGML. 

\begin{table}[tbp]
 \centering
 \caption{
 \ed{Evaluation of} sim-ASIF with micro-F1 score. 
 The training and test sets of these datasets were split using a clustering algorithm such that the distribution of vertex-labels can be largely different.
 }
	\begin{tabular}{cc|rrr}
		\#maxvertices	&Feature\textbackslash Dataset			&AIDS					&Mutagenicity			&NCI1					\\ \hline
		According to \tablename~\ref{tbl:f-score}	&Normal		&0.574$\pm$0.039		&\B{0.720}$\pm$0.014	&0.735$\pm$0.025	\\
		8				&Normal									&0.572$\pm$0.038		&{0.705}$\pm$0.017		&0.726$\pm$0.019	\\
		8				&sim-ASIF \eqref{eq:threshold-feature}	&\B{0.663}$\pm$0.033	&{0.702}$\pm$0.016		&\B{0.755}$\pm$0.017\\
	\end{tabular}
	\label{tbl:k-means-datasplit}
\end{table}

\subsection{Performance on Frequency Feature}

In this section, we evaluate IGML with $g(x)=\log(1+x)$ instead of $g(x)=1_{x>0}$. 
%
Note that because computing the frequency without \ed{overlapping} $\#(H\sqsubseteq G)$ is NP-complete \citep{schreiber2005frequency}, in addition to the exact count, we evaluated the feature defined by an upper bound of $\#(H\sqsubseteq G)$ (see Appendix~\ref{app:freq-approx} for \ed{details}). 
We employed $\log$ \ed{because} the scale of the frequency $x$ is highly diversified.
%
\ed{Based on the results} in Section~\ref{ssec:efficiency}, we \ed{used} \red{WSP}+\red{RSSP} in this section. 
%
The \#maxvertices for each dataset \ed{followed those in} \tablename~\ref{tbl:f-score}. 

The comparison of micro-F1 scores for the exact $\#(H\sqsubseteq G)$ and \ed{approximation} of $\#(H\sqsubseteq G)$ is shown in \tablename~\ref{tbl:approx-log-score}. 
%
The exact $\#(H\sqsubseteq G)$ did not complete five datasets mainly due to the computational difficulty of the frequency counting.
%
In contrast, the approximate $\#(H\sqsubseteq G)$ completed on all datasets. 
Overall, for both the exact and approximate frequency features, the micro-F1 scores were comparable with the case of $g(x)=1_{x>0}$ shown in \tablename~\ref{tbl:f-score}.

\tablename~\ref{tbl:time-freq} \ed{lists} the total \ed{times for} the path-wise optimization for the exact $\#(H\sqsubseteq G)$ and the approximation of $\#(H\sqsubseteq G)$. 
On the AIDS dataset, \ed{the exact} $\#(H\sqsubseteq G)$ did not complete within a day, \ed{while the traversal} time using approximate $\#(H\sqsubseteq G)$ was only 8.6 sec. 
On the BZR dataset, the \ed{traversal} time using the exact $\#(H\sqsubseteq G)$ was seven times that using the approximate $\#(H\sqsubseteq G)$. 
The solving time for the approximation was lower \ed{because} $|\mathcal{F}|$ after traversing of the approximation was significantly less than that of the exact $\#(H\sqsubseteq G)$ in this case.
%
\ed{Because} the approximate $\#(H\sqsubseteq G)$ is an upper bound of the exact $\#(H\sqsubseteq G)$, the variation of the values of the exact $\#(H\sqsubseteq G)$ was smaller than the approximate $\#(H\sqsubseteq G)$.
This resulted in higher correlations among features created by the exact $\#(H\sqsubseteq G)$. 
It is known that the elastic-net regularization tends to select correlated features simultaneously \citep{zou2005regularization}, and therefore, $| \cF |$ in the case of the exact $\#(H\sqsubseteq G)$ becomes larger than in the approximate case. 

\begin{figure}
	\centering
	\includegraphics[width=.5\linewidth]{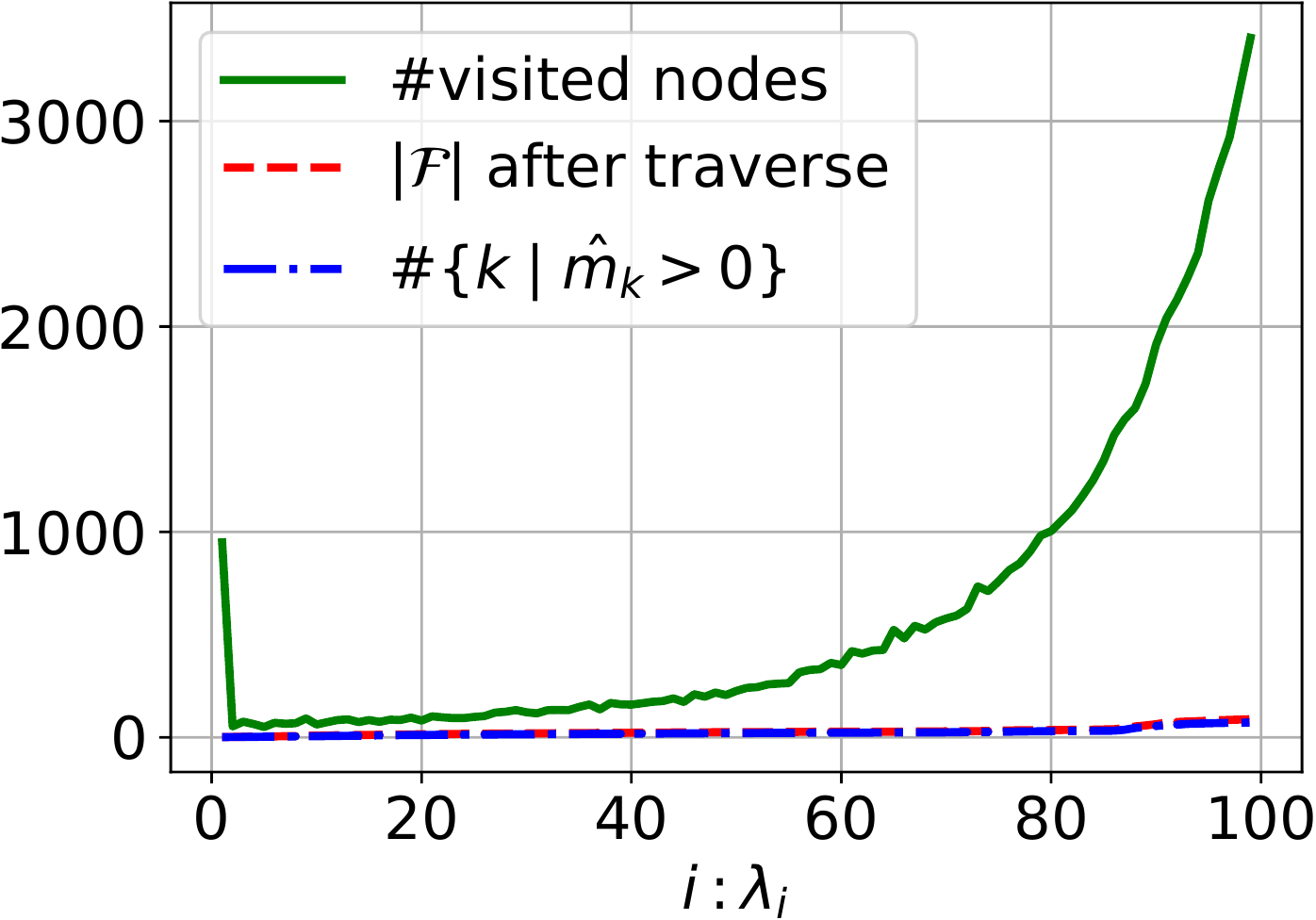}
	\caption{
 \ed{Results} of IGML with $g(x)=\log(1+x)$ on AIDS dataset.
	}
	\label{fig:aids_freq}
\end{figure}

\figurename~\ref{fig:aids_freq} shows the number of visited nodes, size of the feature subset $|\cF|$ after \ed{traversal}, and the number of selected features on the AIDS dataset with the approximate $\#(H\sqsubseteq G)$.
%
This indicates that IGML keeps the number of subgraphs tractable even if $g(x)=\log(1+x)$ is used as the feature. 
%
The \#visited nodes \ed{was less than} 3500, and $|\mathcal{F}|$ after \ed{traversal was} sufficiently close to $|\{k\mid\hat{m}_k>0\}|$. 
%
\red{
We see that \#visited nodes at $\lambda_0$ is larger than many subsequent $\lambda_i$s, and this is the effect of range-based rules, as shown in the case of \figurename~\ref{fig:aids}~\subref{fig:aids_visit}. 
}

\begin{table*}[tbp]
	\caption{
		Micro-F1 \ed{scores for} $g(x)=\log (1+x)$. 
	}
	\centering
	{
	\scriptsize
	\hdashlinewidth=0.2mm
	\hdashlinegap=0.5mm
	\tabcolsep = 3pt
	\begin{tabular}{c|rrrrrrrrrr}
		\multirow{2}{*}{Method\,\textbackslash\,Dataset}	&\multirow{2}{*}{AIDS}		&\multirow{2}{*}{BZR}		&\multirow{2}{*}{DD}			&\multirow{2}{*}{DHFR}		&FRANKE	
										&Mutag		&\multirow{2}{*}{NCI1}		&\multirow{2}{*}{COX2}		&\multirow{2}{*}{DBLP\_v1}	\\
										&			&			&			&			&NSTEIN
										&enicity	&			&			&			
		\\ \hline
		exact						&-			&0.833		&-			&0.802		&-			&-			&-			&0.769		&0.858		\\
		$\#(H\sqsubseteq G)$		&			&$\pm$0.045	&			&$\pm$0.031	&			&			&			&$\pm$0.030	&$\pm$0.005	\\
		\hdashline
		approximation
		&0.982		&0.842		&0.772		&0.791		&0.690		&0.779		&0.762		&0.769		&0.858		\\
		of $\#(H\sqsubseteq G)$
		&$\pm$0.005	&$\pm$0.049	&$\pm$0.026	&$\pm$0.046	&$\pm$0.013	&$\pm$0.010	&$\pm$0.015	&$\pm$0.042	&$\pm$0.005
	\end{tabular}
	}
	\label{tbl:approx-log-score}
\end{table*}

\begin{table}[tbp]
	\centering
	\caption{
 Total time \ed{of} path-wise optimization (sec) \ed{for} $g(x)=\log(1+x)$.
	}
	\label{tbl:time-freq}
	{
	\scriptsize
	\hdashlinewidth=0.2mm
	\hdashlinegap=0.5mm
	\tabcolsep = 3pt
		\begin{tabular}{c|r:r:r|r:r:r}
		Dataset									&\multicolumn{3}{c|}{AIDS}						&\multicolumn{3}{c}{BZR}						\\
		Method~\textbackslash~Process			&Traverse		&Solve			&Total			&Traverse		&Solve			&Total			\\ \hline 
		exact $\#(H\sqsubseteq G)$				&\multicolumn{3}{c|}{$>$ a day}&1662.2$\pm$93.0&93.0$\pm$19.4	&1755.2$\pm$213.5	\\
		approximation of $\#(H\sqsubseteq G)$	&{8.6}$\pm$1.4	&14.5$\pm$1.4	&{23.1}$\pm$1.9	&236.0$\pm$26.1	&13.0$\pm$~3.1	&249.0$\pm$~28.9	\\
		\end{tabular}
	}
\end{table}

\section{Conclusions}

\ed{In this paper, we} proposed an interpretable metric learning method for graph data, named \emph{interpretable graph metric learning} (IGML).
%
To avoid computational difficulty, we \ed{built} an optimization algorithm that combines safe screening, working set selection, and their pruning extensions.
We also discussed the three extensions of IGML\ed{:} (a) applications to other structured data, (b) triplet loss-based formulation, and (c) incorporating vertex-label similarity into the feature.
We empirically evaluated the performance of IGML compared with existing graph classification methods.
Although IGML was the only method \ed{with} clear interpretability, it showed superior or comparable prediction performance compared to other state-of-the-art methods.
%
\ed{The} practicality of IGML was \ed{further demonstrated} through some illustrative examples of identified subgraphs.
\red{
Although IGML optimized the metric within tractable time in the experiments, the subgraphs were restricted to moderate sizes (up to $30$), and a current major bottleneck for extracting larger-sized subgraphs is the memory requirement of the gSpan tree.
Therefore, mitigating this memory consumption is an important future directions to apply IGML to a wider class of problems.
}


%
%

\section*{Funding}
This work was supported by MEXT KAKENHI to I.T. (16H06538, 17H00758) and M.K. (16H06538, 17H04694); from JST CREST awarded to I.T. (JPMJCR1302, JPMJCR1502) and PRESTO awarded to M.K. (JPMJPR15N2); from the Collaborative Research Program of Institute for Chemical Research, Kyoto University to M.K. (grant \#2018-33 and \#2021-31); from the MI2I project of the Support Program for Starting Up Innovation Hub from JST awarded to I.T., and M.K.; and from RIKEN Center for AIP awarded to I.T.

\section*{Conflicts of interest/Competing interests}
\ed{The authors declare no conflicts of interest.}

\section*{Availability of data and material}
All datasets used in the experiments are available on \ed{online} (\ed{see} Section~\ref{sec:experiments} for \ed{details}). 

\section*{Code availability}
The source code \ed{for} the program used in the experiments is available at 
\url{https://github.com/takeuchi-lab/Learning-Interpretable-Metric-between-Graphs}. 


\bibliographystyle{apalike}
\bibliography{ref}

\clearpage
\appendix

\noindent
{\Large {\bf 
Appendix
}
}

\section{Dual Problem}
\label{app:dual}
The primal problem \eqref{eq:primal} can be re-written as
\[
\begin{split}
 \min_{\bm m, \bm z}~&\sum_{i\in[n]}[\sum_{l\in \mathcal{D}_i}\ell_L(z_{il})+\sum_{j\in \mathcal{S}_i}\ell_{-U}(z_{ij})]
 +\lambda R(\bm m)\\
 \mathrm{s.t.}~&\bm m\ge\bm 0, z_{il}=\bm m^\top \bm c_{il}, \ z_{ij}=-\bm m^\top \bm c_{ij}.
\end{split}
\]
The Lagrange function $\cL$ is
\begin{multline*}
 \cL(\bm m, \bm z, \bm\alpha, \bm\beta)
 \coloneqq\sum_{i\in[n]}[\sum_{l\in \mathcal{D}_i}\ell_L(z_{il})+\sum_{j\in \mathcal{S}_i}\ell_{-U}(z_{ij})]
 +\lambda R(\bm m)\\
 +\sum_{i\in[n]}[\sum_{l\in \mathcal{D}_i}\alpha_{il}(z_{il}-\bm m^\top\bm c_{il})+\sum_{j\in \mathcal{S}_i}\alpha_{ij}(z_{ij}+\bm m^\top\bm c_{ij})]-\bm\beta^\top\bm m,
\end{multline*}
where $\bm\alpha\in\mathbb{R}^{2nK}$ and $\bm\beta\in\mathbb{R}_+^p$ are Lagrange multipliers.
The dual function $D_\lambda$ is then
\begin{equation}
	\label{eq:dual_function}
	D_\lambda(\bm \alpha,\bm \beta)\coloneqq\inf_{\bm m,\bm z}\cL(\bm m, \bm z, \bm\alpha, \bm\beta).
\end{equation}
\ed{By} the definition of \ed{the} dual function \ed{in} \eqref{eq:dual_function}, \ed{to} minimize $\mathcal{L}$ with respect to $\bm m$, by partially differentiating $\mathcal{L}$, we obtain
\begin{equation}\label{eq:nablam}
	\nabla_{\bm m}\cL=\lambda(\bm 1+\eta\bm m)+
	\sum_{i\in[n]}[-\sum_{l\in \mathcal{D}_i}\alpha_{il}\bm c_{il}+\sum_{j\in \mathcal{S}_i}\alpha_{ij}\bm c_{ij}]-\bm\beta=\bm0. 
\end{equation}
%
The convex conjugate function of $\ell_t$ is
\begin{equation}\label{eq:conj}
	\ell_t^*(-\alpha_{ij})=\sup_{z_{ij}} \{(-\alpha_{ij})z_{ij}-\ell_t(z_{ij})\}, 
\end{equation}
which can be written as 
\begin{equation}\label{eq:conj-2}
\ell_t^*(x_*)=\frac{1}{4}x_*^2+tx_*, (x_*\le0). 
\end{equation}
%
\ed{From} \eqref{eq:nablam}, \eqref{eq:conj}, and \eqref{eq:conj-2}, the dual function can be written as 
\begin{align*}
	&D_\lambda(\bm \alpha,\bm \beta)\\
	&=
	-\sum_{i\in[n]}[\sum_{l\in \mathcal{D}_i} \ell_L^*(-\alpha_{il})+\sum_{j\in \mathcal{S}_i}\ell_{-U}^*(-\alpha_{ij})]
	-\frac{\lambda\eta}{2}\|\bm m_{\lambda} (\bm\alpha, \bm\beta)\|_2^2\\
	&=-\frac{1}{4}\|\bm\alpha\|_2^2+\bm t^\top\bm\alpha
	-\frac{\lambda\eta}{2}\|\bm m_{\lambda} (\bm\alpha, \bm\beta)\|_2^2.
\end{align*}
where 
\begin{align*}
	\bm m_\lambda(\bm\alpha,\bm\beta)&\coloneqq 
	\frac{1}{\lambda\eta}\left[\bm\beta+
	\sum_{i\in[n]}(\sum_{l\in \mathcal{D}_i}\alpha_{il}\bm c_{il}-\sum_{j\in \mathcal{S}_i}\alpha_{ij}\bm c_{ij})-\lambda\bm1
	\right]\\
	&=
	\frac{1}{\lambda\eta}[\bm\beta+\bm C\bm \alpha-\lambda\bm 1]. 
\end{align*}
Therefore, although the dual problem can be written as 
\[
	\max_{\bm\alpha\ge\bm0, \bm\beta\ge\bm 0} D(\bm\alpha,\bm\beta),
\]
by maximizing $D(\bm\alpha,\bm\beta)$ with respect to $\bm\beta$, we obtain a more straightforward dual problem \eqref{eq:dual}. 

We obtain $\alpha_{ij}=-\ell'_t(z_{ij})$, used in \eqref{eq:primal2dual}, from the derivative of $\cL$ with respect to $z_{ij}$. 

\section{Proof of Lemma~\ref{lmm:pruning}}
\label{app:pruning}
\ed{From} \eqref{eq:monotonicity}, 
the value of $(x_{i,k'}-x_{j,k'})^2$ is bounded as follows: 
\begin{align*}
(x_{i,k'}-x_{j,k'})^2
&\le \max_{0\le x_{i,k'}\le x_{i,k}, 0\le x_{j,k'}\le x_{j,k}} (x_{i,k'}-x_{j,k'})^2\\
&= \max\{x_{i,k}, x_{j,k}\}^2. 
\end{align*}
Using this inequality, the inner product $\bm C_{k',:}\bm q$ is likewise bounded: 
\begin{align*}
	\bm C_{k',:}\bm q
	&=\sum_{i\in[n]}\Bigl[
	\sum_{l\in \mathcal{D}_i}q_{il}(x_{i,k'}-x_{l,k'})^2-\sum_{j\in \mathcal{S}_i}q_{ij}(x_{i,k'}-x_{j,k'})^2
	\Bigr]\\
	&\le 
	\sum_{i\in[n]}\sum_{l\in \mathcal{D}_i}q_{il}\max\{x_{i,k},x_{l,k}\}^2. 
\end{align*}
Similarly, the norm $\|\bm C_{k',:}\|_2$ is bounded: 
\begin{align*}
	\|\bm C_{k',:}\|_2
	&=\sqrt{\sum_{i\in[n]}[\sum_{l\in\mathcal{D}_i}(x_{i,k'}-x_{l,k'})^4+\sum_{j\in\mathcal{S}_i}(x_{i,k'}-x_{j,k'})^4]}\\
	&\le\sqrt{\sum_{i\in[n]}[\sum_{l\in\mathcal{D}_i}\max\{x_{i,k},x_{l,k}\}^4+\sum_{j\in\mathcal{S}_i}\max\{x_{i,k},x_{j,k}\}^4]}. 
\end{align*}
Therefore, $\bm C_{k',:}\bm q+r\|\bm C_{k',;}\|_2 $ is bounded by $\mathrm{Prune}(k | \bm q, r)$. 

\section{Proof of Lemma~\ref{lmm:pruning-binary}}
\label{app:}

First, we consider the first term of $\*C_{k',:} \*q + r \| \*C_{k',:}\|_2$:
\[
\bm C_{k',:}\bm q=\sum_{i\in[n]}\Bigl[
\underbrace{\sum_{l\in \mathcal{D}_i}q_{il}(x_{i,k'}-x_{l,k'})^2-\sum_{j\in \mathcal{S}_i}q_{ij}(x_{i,k'}-x_{j,k'})^2}_{\coloneqq {\rm diff}}
\Bigr].
\]
%
%
Now, $x_{i,k'}\in\{0,1\}$ is assumed.
Then, if $x_{i,k'}=0$, we obtain
\[
{\rm diff} = \sum_{l\in \mathcal{D}_i}q_{il}x_{l,k'}-\sum_{j\in \mathcal{S}_i}q_{ij}x_{j,k'}\le\sum_{l\in \mathcal{D}_i}q_{il}x_{l,k}. 
\]
\ed{Meanwhile}, if $x_{i,k'}=1$, we \ed{have} $x_{i,k}=1$ from the monotonicity, and subsequently
{\small\[
{\rm diff} = \sum_{l\in \mathcal{D}_i}q_{il}(1-x_{l,k'})-\sum_{j\in \mathcal{S}_i}q_{ij}(1-x_{j,k'})
	\le
		\sum_{l\in \mathcal{D}_i}q_{il}-\sum_{j\in \mathcal{S}_i}q_{ij}(1-x_{j,k}).
\]}
By using ``$\max$'', we can unify these two upper bounds into 
	{\small\[
		\bm C_{k',:}\bm q\le \sum_{i\in[n]}\max\Bigl\{\sum_{l\in \mathcal{D}_i}q_{il}x_{l,k} 
		\ , \  
		x_{i,k}[\sum_{l\in \mathcal{D}_i}q_{il}-\sum_{j\in \mathcal{S}_i}q_{ij}(1-x_{j,k})]\Bigr\}. 
	\]}
Employing a similar concept, the norm of $\*C_{k',:}$ can also be bounded by
	{\small\begin{align*}
		\|\bm C_{k',:}\|_2
		&=\sqrt{\sum_{i\in[n]}[\sum_{l\in\mathcal{D}_i}(x_{i,k'}-x_{l,k'})^4+\sum_{j\in\mathcal{S}_i}(x_{i,k'}-x_{j,k'})^4]}\\
		&\le\sqrt{\sum_{i\in[n]}[\sum_{l\in\mathcal{D}_i}\max\{x_{i,k},x_{l,k}\}+\sum_{j\in\mathcal{S}_i}\max\{x_{i,k},x_{j,k}\}]}.
	\end{align*}}
Thus, we obtain
	{\small\begin{multline*}
		\mathrm{Prune}(k)\!\coloneqq\!\sum_{i\in[n]}\max\{\sum_{l\in \mathcal{D}_i}q_{il}x_{l,k}, x_{i,k}[\sum_{l\in \mathcal{D}_i}q_{il}-\sum_{j\in \mathcal{S}_i}q_{ij}(1-x_{j,k})]\}\\
		+r\sqrt{\sum_{i\in[n]}[\sum_{l\in\mathcal{D}_i}\max\{x_{i,k},x_{l,k}\}+\sum_{j\in\mathcal{S}_i}\max\{x_{i,k},x_{j,k}\}]}.
	\end{multline*}}

\section{Proof of Theorem~\ref{thm:DGB} (DGB)}
\label{app:DGB}

 From \ed{the} $1/2$-strong convexity of $-D_\lambda(\bm \alpha)$, for any $\bm\alpha\ge\bm0$ and $\bm\alpha^\star\ge\bm0$, we obtain
 \[
  D_\lambda(\bm\alpha)\le 
  D_\lambda(\bm\alpha^\star)+\nabla D_\lambda(\bm\alpha^\star)^\top (\bm\alpha-\bm\alpha^\star)
  -\frac{1}{4}\|\bm\alpha-\bm\alpha^\star\|_2^2.
  \addtag \label{eq:strong_convexity}
 \]
 Applying weak duality 
 $P_\lambda(\bm m)\ge D_\lambda(\bm\alpha^\star)$
 and the optimality condition of the dual problem
 $\nabla D_\lambda(\bm\alpha^\star)^\top (\bm\alpha-\bm\alpha^\star)\le 0$
 to \eqref{eq:strong_convexity}, we obtain DGB.

\section{Proof of Theorem~\ref{thm:RPB} (RPB)}
\label{app:RPB}
From the optimality condition of the dual problem \eqref{eq:dual}, 
\begin{align} 
 \nabla_{\bm\alpha} D_{\lambda_0}(\bm \alpha_0^\star)^\top
 (\frac{\lambda_0}{\lambda_1}\bm \alpha_1^\star-\bm \alpha_0^\star)
 \le0, 
 \label{eq:for_dual_opt1}
 \\
 \nabla_{\bm\alpha} D_{\lambda_1}(\bm \alpha_1^\star)^\top
 (\frac{\lambda_1}{\lambda_0}\bm \alpha_0^\star-\bm \alpha_1^\star)
 \le0. 
 \label{eq:for_dual_opt2}
\end{align}
Here, the gradient vector \ed{for} the optimal solution is 
\begin{align*}
	\nabla D_{\lambda_i}(\bm\alpha_i^\star)
	&=-\frac{1}{2}\bm\alpha_i^\star+\bm t-\bm C^\top \bm m_{\lambda_i}(\bm\alpha_i^\star)\\
	&=-\frac{1}{2}\bm\alpha_i^\star+\bm t-\bm C^\top{\bm m_i^\star}\ed{.}
\end{align*}
\ed{Thus}, by substituting this equation into \eqref{eq:for_dual_opt1} and \eqref{eq:for_dual_opt2}, \ed{we get}
\begin{gather}
	\label{eq:dual_opt1}
	(-\frac{1}{2}\bm\alpha_0^\star+\bm t-\bm C^\top{\bm m_0^\star})^\top(\frac{\lambda_0}{\lambda_1}\bm \alpha_1^\star-\bm \alpha_0^\star)\le0, \\
	\label{eq:dual_opt2}
	(-\frac{1}{2}\bm\alpha_1^\star+\bm t-\bm C^\top{\bm m_1^\star})^\top(\frac{\lambda_1}{\lambda_0}\bm \alpha_0^\star-\bm \alpha_1^\star)\le0. 
\end{gather}
From $\lambda_1 \times \eqref{eq:dual_opt1}+\lambda_0 \times \eqref{eq:dual_opt2}$, 
\begin{equation}\label{eq:dual_opt_add}
	(-\frac{1}{2}[\bm\alpha_0^\star-\bm\alpha_1^\star]
	-\bm C^\top[\bm m_0^\star-\bm m_1^\star])^\top
	(\lambda_0\bm \alpha_1^\star-\lambda_1\bm \alpha_0^\star)\le0. 
\end{equation}
\ed{From} \eqref{eq:nablam}, 
\begin{equation}\label{eq:calpha}
	\bm C\bm \alpha_i=\lambda_i\eta\bm m_i+\lambda_i\bm1-\bm\beta_i.
\end{equation}
By substituting equation \eqref{eq:calpha} into equation \eqref{eq:dual_opt_add}, \ed{we get}
\[
	-\frac{1}{2}[\bm\alpha_0^\star-\bm\alpha_1^\star]^\top(\lambda_0\bm \alpha_1^\star-\lambda_1\bm \alpha_0^\star)
	-[\bm m_0^\star-\bm m_1^\star]^\top(\lambda_0\lambda_1\eta[\bm m_1-\bm m_0]-\lambda_0\bm\beta_1^\star+\lambda_1\bm\beta_0^\star )\le0.
\]
Transforming this inequality \ed{by} completing the square
with the complementary \ed{conditions} ${\bm m_i^\star}^\top\bm\beta_i^\star=0$ and ${\bm m_1^\star}^\top\bm\beta_0^\star,{\bm m_0^\star}^\top\bm\beta_1^\star\ge0$, we obtain
\[
	\left\|\bm \alpha_1^\star-\frac{\lambda_0+\lambda_1}{2\lambda_0}\bm \alpha_0^\star\right\|_2^2
	+2\lambda_1\eta\|\bm m_0^\star - \bm m_1^\star\|_2^2
	\le \left\|\frac{\lambda_0-\lambda_1}{2\lambda_0}\bm \alpha_0^\star\right\|_2^2.
\]
\ed{Applying}
$\|\bm m_0^\star - \bm m_1^\star\|_2^2\ge0$ to this inequality, we obtain RPB.

\section{Proof of Theorem~\ref{thm:RRPB} (RRPB)}
\label{app:RRPB}
Considering a hypersphere that expands the RPB radius by $\frac{\lambda_0+\lambda_1}{2\lambda_0}\epsilon$ 
and replaces the RPB center with $\frac{\lambda_0+\lambda_1}{2\lambda_0}\bm{\alpha}_0$, 
we obtain 
\[
\left\|\bm{\alpha}_1^\star-\frac{\lambda_0+\lambda_1}{2\lambda_0}\bm{\alpha}_0\right\|_2
\le
\frac{|\lambda_0-\lambda_1|}{2\lambda_0}\left\|\bm{\alpha}_0^\star\right\|_2
+\frac{\lambda_0+\lambda_1}{2\lambda_0}\epsilon.
\]
\ed{Because}
$\epsilon$ is defined by
$\|\bm{\alpha}_0^\star-\bm{\alpha}_0\|_2\le\epsilon$,
this sphere covers any RPB made by $\bm{\alpha}_0^\star$ which satisfies $\|\bm{\alpha}_0^\star-\bm{\alpha}_0\|_2\le\epsilon$. 
Using the reverse triangle inequality
\[
\|\bm{\alpha}_0^\star\|_2-\|\bm{\alpha}_0\|_2\le\|\bm{\alpha}_0^\star-\bm{\alpha}_0\|_2\le\epsilon,
\]
the following is obtained.
\[
\left\|\bm{\alpha}_1^\star-\frac{\lambda_0+\lambda_1}{2\lambda_0}\bm{\alpha}_0\right\|_2
\le
\frac{|\lambda_0-\lambda_1|}{2\lambda_0}(\left\|\bm{\alpha}_0\right\|_2+\epsilon)
+\frac{\lambda_0+\lambda_1}{2\lambda_0}\epsilon.
\]
By \ed{rearranging} this, RRPB is obtained.

\section{Proof for Theorem~\ref{thm:RSS} (RSS), \ref{thm:RSP} (RSP) and \ref{thm:RSP-bin} (RSP for binary feature)}
\label{app:}

%

\ed{Here, we address only Theorems}~\ref{thm:RSS} and \ref{thm:RSP} because \ed{Theorem}~\ref{thm:RSP-bin} can be derived in almost the same way as \ed{Theorem}~\ref{thm:RSP}.
When $\lambda_1 = \lambda$ is set in RRPB, the center and the radius of the bound
$\cB = \{ \*\alpha \mid \| \*\alpha - \*q \|_2^2 \leq r^2 \}$
are 
$\*q = \frac{\lambda_0+\lambda}{2\lambda_0}\bm \alpha_0$
and
$r = \left\|\frac{\lambda_0-\lambda}{2\lambda_0}\bm \alpha_0\right\|_2+\Bigl(\frac{\lambda_0+\lambda}{2\lambda_0}+\frac{|\lambda_0-\lambda|}{2\lambda_0}\Bigr)\epsilon$, \ed{respectively}.
Substituting these $\*q$ and $r$ into \eqref{eq:screeningRule} and \eqref{eq:pruningRule}, respectively, and \ed{rearranging} them, we can obtain the range in which \ed{the} screening and pruning conditions hold.

\section{Proof of Theorem~\ref{thm:WS} (Convergence of WS)}
\label{app:WS}
By introducing a new variable $\bm s$, 
the dual problem \eqref{eq:dual} can be written as 
\[
	\begin{split}
		\max_{\bm \alpha\ge\bm 0, \bm s\ge\bm 0}	&~-\frac{1}{4}\|\bm \alpha\|^2+\bm t^\top\bm \alpha-\frac{1}{2\lambda\eta}\|\bm s\|^2\\
		\mathrm{s.t.}				&~\bm C\bm \alpha-\lambda\bm 1-\bm s\le\bm 0.
	\end{split}
\]
%
We demonstrate the convergence of \ed{the WS} method on a more general convex problem as follows: 
\begin{equation}
	\bm x^\star\coloneqq \mathrm{arg}\min_{\bm x\in \mathcal{D}}f(\bm x)~\mathrm{s.t.}~h_i(\bm x)\le 0, \forall i\in[n], 
	\label{eq:originalWSproblem}
\end{equation}
where $f(\bm x)$ is a $\gamma$-strong convex function ($\gamma>0$). 
%
%
Here, as shown in Algorithm~\ref{alg:workingSetMethod}, the working set is defined by $\mathcal{W}_t = \{ j \mid h_j(\bm x_{t-1}) \ge 0 \}$ at every iteration.
Then, the updated working set includes all the violated constraints and the constraints on the boundary.
We show that Algorithm~\ref{alg:workingSetMethod} finishes with finite $T$-steps and returns the optimal solution $\bm x_{T}=\bm x^\star$. 

\begin{algorithm}[tbp]
 \caption{General \ed{WS} Method}
	\label{alg:workingSetMethod}
	initialize $\bm x_0\in\mathcal{D}$\;
	\For{$t=1,2,... ~\mathbf{until}~\mathrm{converged}$}{
                $\mathcal{W}_t = \{ j \mid h_j(\bm x_{t-1}) \ge 0 \}$\;
		$\bm x_t=\mathrm{arg}\min_{\bm x\in\mathcal{D} }f(\bm x)~\mathrm{s.t.}~h_j(\bm x)\le0, \forall j\in\mathcal{W}_t$\;
	}
\end{algorithm}

\begin{proof}
 \ed{Because} $f$ is $\gamma$-strong convex from the assumption, the following inequality holds\ed{:} 
	\begin{equation}
		\label{eq:strong-convex}
		f(\bm x_{t+1})\ge f(\bm x_{t})+\nabla f(\bm x_{t})^\top (\bm x_{t+1}-\bm x_{t})+\frac{\gamma}{2}\|\bm x_{t+1}-\bm x_{t}\|^2. 
	\end{equation}
 At step $t$, the problem can be written \ed{using} only the active constraint at the optimal solution $\bm x_t$ \ed{as follows:}
	\begin{align}
	\bm x_t	&=\mathrm{arg}\min_{\bm x\in\mathcal{D}}f(\bm x)~\mathrm{s.t.}~h_i(\bm x)\le 0,\forall i\in\mathcal{W}_t \nonumber\\
			&=\mathrm{arg}\min_{\bm x\in\mathcal{D}}f(\bm x)~\mathrm{s.t.}~h_i(\bm x)\le 0,\forall i\in\{j\in\mathcal{W}_t\mid h_j(\bm x_t)=0\}
			\label{eq:step-t-rewitten}
	\end{align}
 From the definition of $\cW_t$, the working set $\cW_{t+1}$ must contain all active constraints $\{ j \in \cW_t \mid h_j(\*x_t) = 0\}$ at the step $t$ and can contain other constraints that are not included in $\mathcal{W}_t$.
 This means that $\bm x_{t+1}$ must be in the feasible region of the optimization problem at the step $t$ \eqref{eq:step-t-rewitten}:
	\[
	\mathcal{F}\coloneqq\left\{\bm x\in\mathcal{D}\mid h_i(\bm x)\le 0,\forall i\in\{j\in\mathcal{W}_t\mid h_j(\bm x_t)=0\}\right\}
	\] 
	Therefore, from the optimality condition of the optimization problem \eqref{eq:step-t-rewitten}, 
	\begin{equation}
		\label{eq:optimality}
		\nabla f(\bm x_t)^\top (\bm x_{t+1}-\bm x_t)\ge0, \bm x_{t+1}\in\mathcal{F}. 
	\end{equation}
 From the \ed{inequalities} \eqref{eq:strong-convex} \ed{and} \eqref{eq:optimality}, we obtain
	\[
	f(\bm x_{t+1})\ge f(\bm x_t)+\frac{\gamma}{2}\|\bm x_{t+1}-\bm x_t\|^2.
	\]
 If $\*x_t$ is not optimal, there exists at least one violated constraint $h_{j'}(\*x_t) > 0$ for some $j'$ because otherwise $\*x_t$ is optimal.
 Then, we see $\*x_{t+1} \neq \*x_t$ because $\*x_{t+1}$ should satisfy the constraint $h_{j'}(\*x_{t+1}) \leq 0$.
 If $\bm x_t\ne \bm x_{t+1}$, \ed{from} $\|\bm x_{t+1}-\bm x_t\|^2>0$, 
	\[
	f(\bm x_{t+1})\ge f(\bm x_t)+\frac{\gamma}{2}\|\bm x_{t+1}-\bm x_t\|^2>f(\bm x_t). 
	\]
 Thus, the objective function always strictly increases ($f(\bm x_t)<f(\bm x_{t+1})$). 
 This indicates that the algorithm never encounters the same working set $\cW_t$ as the set of other iterations $t' \neq t$.
 For any step $t$, the optimal value $f(\bm x_t)$ with a subset of the original constraints $\mathcal{W}_t$ must be smaller than or equal to the optimal value $f(\bm x^\star)$ of original problem \eqref{eq:originalWSproblem} with all constraints. 
 Therefore, $f(\bm x_t)\le f(\bm x^\star)$ is satisfied, and we obtain $f(\bm x_T)=f(\bm x^\star)$ at some finite step $T$. 
\end{proof}

\section{CPU Time for Other \ed{Datasets}}
\label{app:cpu-time}

\tablename~\ref{tbl:time-appendix} \ed{lists the} computational \ed{times on} the BZR, DD, and FRANKENSTEIN datasets.
%
We first note that \red{RSSP} was \ed{approximately} 2--4 times faster in terms of the \ed{traversal} time compared with \red{SSP}.
Next, comparing \red{RSSP} and \red{WSP}, we see that \red{RSSP} was faster for Traverse, and \red{WSP} was faster for Solve, as we \ed{observed} in \tablename~\ref{tbl:time}.
Thus, the combination of WS\&SP and \red{RSSP} \ed{was} the fastest for all three datasets in total.

\begin{table*}[tbp]
	\centering
	\caption{Total \ed{times for} the path-wise optimization (sec). }
	\label{tbl:time-appendix}
	{
	\scriptsize
	\hdashlinewidth=0.2mm
	\hdashlinegap=0.5mm
	\tabcolsep=3pt
		\begin{tabular}{c|r:r:r|r:r:r|r:r:r}
		Dataset	&\multicolumn{3}{c|}{BZR}			&\multicolumn{3}{c|}{DD}			&\multicolumn{3}{c}{FRANKENSTEIN}	\\
		Method\,\textbackslash\,Process	
				&Traverse	&Solve		&Total		&Traverse	&Solve		&Total		&Traverse	&Solve		&Total		\\
				\hline
		\red{SSP}	&1397.1		&			&4281.9		&4292.9		&			&13961.3	&249.1		&			&5013.0		\\
					&$\pm$91.7	&2884.8		&$\pm$964.1	&$\pm$388.3	&9668.4		&$\pm$1580.6&$\pm$9.3	&4763.9		&$\pm$442.4	\\
				\cdashline{1-2}\cdashline{4-5}\cdashline{7-8}\cdashline{10-10}
		\red{RSSP}		&\B{539.2}	&$\pm$934.5	&3424.0		&1132.2		&$\pm$1267.5&10800.6	&\B{189.7}	&$\pm$441.5	&4953.6		\\
				&$\pm$47.2	&			&$\pm$956.9	&$\pm$118.0	&			&$\pm$1354.9&$\pm$8.4	&			&$\pm$439.1	\\
				\hline
		\red{WSP}	&2448.5		&			&2724.3		&5888.3		&			&7652.8		&380.1		&			&938.3		\\
				&$\pm$170.8	&\B{275.8}	&$\pm$184.9	&$\pm$465.6	&\B{1764.5}	&$\pm$622.6	&$\pm$12.4	&\B{558.2}	&$\pm$57.5	\\
				\cdashline{1-2}\cdashline{4-5}\cdashline{7-8}\cdashline{10-10}
		\red{WSP}+	&565.5		&$\pm$68.5	&\B{841.3}	&\B{946.1}	&$\pm$195.6	&\B{2710.6}	&233.0		&$\pm$56.5	&\B{791.1}	\\
		\red{RSSP} &$\pm$49.7	&			&$\pm$97.3	&$\pm$83.1	&			&$\pm$258.6	&$\pm$11.7	&			&$\pm$55.7	\\
		\end{tabular}
	}
\end{table*}

\section{Approximating Frequency Without Overlap}
\label{app:freq-approx}

\red{
Let $F_G(H)$ be ``frequency without overlap'' that is the frequency of a subgraph of a given graph where any shared vertices and edges are disallowed for counting.
This $F_G(H)$ is non-increasing with respect to the growth of $H$, but computing it is computationally complicated. 
}
\ed{Assuming} that we know where all the subgraphs $H$ appear in graph $G$, calculating \red{$F_G(H)$}
is equivalent to the problem of finding the maximum independent set, \ed{which} is NP-complete \citep{schreiber2005frequency}. 
In this section, using information obtained in the process of generating \ed{the} gSpan tree, we approximate the frequency without overlap by its upper bound. 
\red{
This upper bound is also a lower bound of the frequency with overlap.
}

\begin{figure}[tb]
	\centering
	\includegraphics[width=0.75\linewidth]{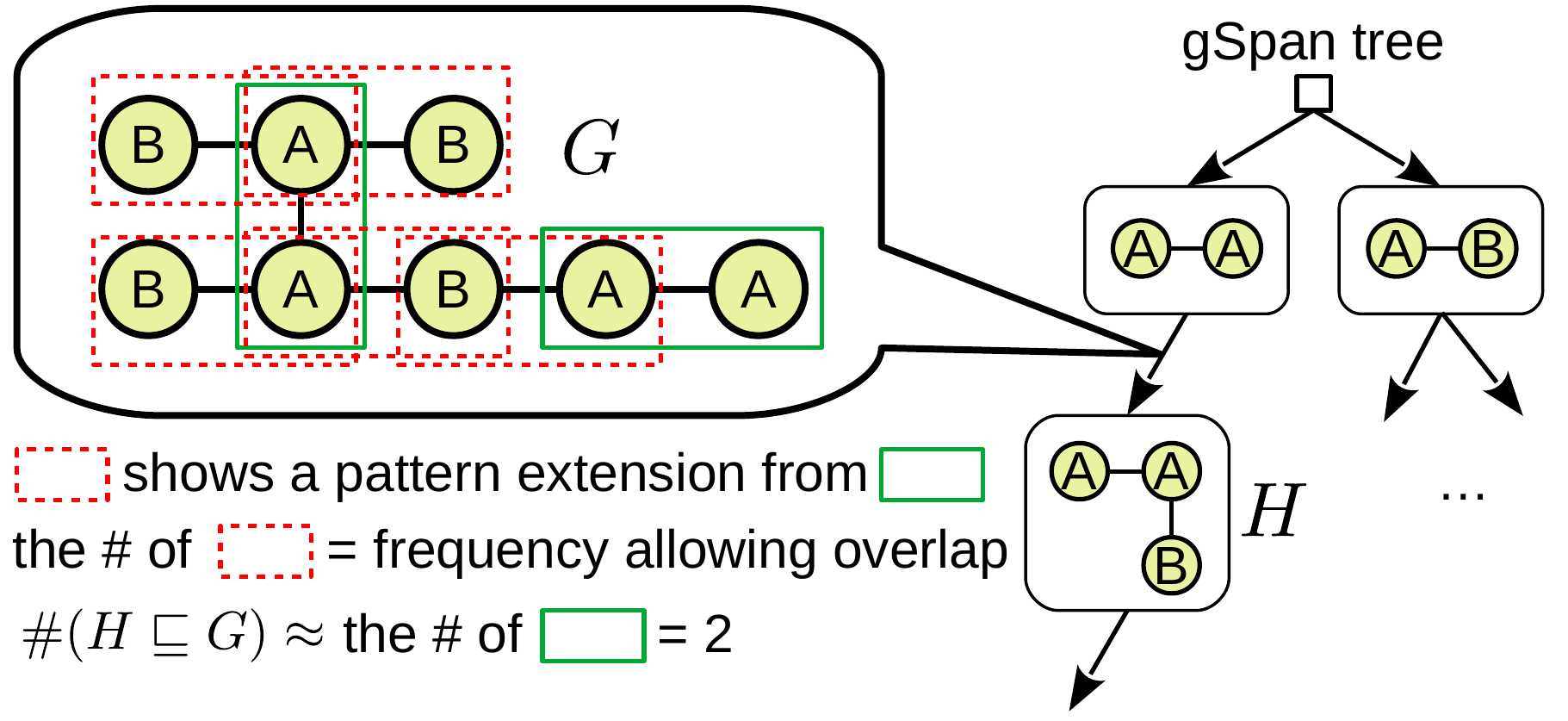}
	\caption{Approximation of $\#(H\sqsubseteq G)$. }
	\label{fig:frequencyApprox}
\end{figure}

\figurename~\ref{fig:frequencyApprox} shows the process of generating the gSpan tree and \ed{frequency}. 
In the figure, we consider the frequency of the subgraph $H$ (\textcircled{\scriptsize A}-\textcircled{\scriptsize A}-\textcircled{\scriptsize B}) contained in the graph $G$. 
The graph $H$ is obtained as a pattern extension of graph \textcircled{\scriptsize A}-\textcircled{\scriptsize A} (green frame) by \textcircled{\scriptsize A}-\textcircled{\scriptsize B} (red frame). 
%
%
%
%
%
%
%
\red{
gSpan stores the number of these pattern extensions at each traverse node. 
We define the count by this extension as $F_G^{\rm max}(H)$ (e.g., $F_G^{\rm max}(H) = 5$ for \textcircled{\scriptsize A}-\textcircled{\scriptsize A}-\textcircled{\scriptsize B}).
Note that $F_G^{\rm max}(H)$ is the frequency of $H$ allowing overlap and duplicately counting matches that are equivalent except for the index of nodes (e.g., $F^{\rm max}_G(H)$ for 
\textcircled{\scriptsize A}-\textcircled{\scriptsize A}-\textcircled{\scriptsize B}-\textcircled{\scriptsize A}-\textcircled{\scriptsize A}
is two in the figure).
Suppose that $H$ currently has $e (>1)$ edges (for example, $e = 2$ in \textcircled{\scriptsize A}-\textcircled{\scriptsize A}-\textcircled{\scriptsize B}).
We recursively go back the traverse tree (a tree in the right of \figurename~\ref{fig:frequencyApprox}) until we reach $e = 1$, i.e., the starting edge that generates $H$ (in the case of \textcircled{\scriptsize A}-\textcircled{\scriptsize A}-\textcircled{\scriptsize B}, the starting edge is \textcircled{\scriptsize A}-\textcircled{\scriptsize A}).
We use the number of unique matches of this starting edges (the number of green frames), which we define as $F^{\rm approx}_G(H)$, as an approximation of $F_G(H)$.
%
Obviously, $F^{\rm approx}_G(H)$ is less than or equal to $F_G^{\rm max}(H)$.
%
In the example, the number of {green frames} must be less than or equal to the number of {red frames} .
%
Further, {because} only overlaps on the starting edge $e = 1$ {are} considered instead of overlaps in entire $H$, $F^{\rm approx}_G(H)$ is greater than or equals to $F_G(H)$.
%
Therefore, overall, we have $F_G(H) \leq F^{\rm approx}_G(H) \leq F_G^{\rm max}(H)$.
Unfortunately, from the definition,
$F^{\rm approx}_G(H)$
gives the same value whenever $H$ has the same starting edge.
%
However, this means that $F^{\rm approx}_G(H)$ satisfies the monotonicity constraint for our pruning. 
%
Because the subgraph counting is a difficult problem and is not the main focus of our study, we employ $F^{\rm approx}_G(H)$ as a simple approximation.
For our framework, any approximation is applicable given that it satisfies the monotonicity constraint.
}

\end{document}